\documentclass[onecolumn]{IEEEtran}
\usepackage[noadjust]{cite}

\ifCLASSINFOpdf
   \usepackage[pdftex]{graphicx}
\else
 
   \usepackage[dvips]{graphicx}
\fi

\usepackage{amsmath}

\usepackage{MnSymbol}
\usepackage{wasysym}
\usepackage{tikz}
\newcommand{\newsymbol}{
\begin{tikzpicture}
\filldraw[fill=black,draw=black] circle (2.5pt);
\end{tikzpicture}
}

\usepackage[caption=false,font=footnotesize]{subfig}

\DeclareMathOperator*{\argmin}{\arg\!\min}

\begin{document}

\title{On the Fusion of Compton Scatter and Attenuation Data for Limited-view X-ray Tomographic Applications}%
 \author{Hamideh Rezaee, Brian Tracey, Eric Miller\\hamideh.rezaee, brian.tracey, eric.miller @tufts.edu\\Department of Electrical and Computer Engineering, Tufts University, Medford, MA, USA}
 

\maketitle

\begin{abstract}
 In this paper we demonstrate the utility of fusing energy-resolved observations of Compton scattered photons with traditional attenuation data for the joint recovery of mass density and photoelectric absorption in the context of limited view tomographic imaging applications. We begin with the development of a physical and associated numerical model for the Compton scatter process. Using this model, we propose a variational approach recovering these two material properties. In addition to the typical data-fidelity terms, the optimization functional includes regularization for both the mass density and photoelectric coefficients. We consider a novel edge-preserving method in the case of mass density. To aid in the recovery of the photoelectric information, we draw on our recent method in \cite{r15} and employ a non-local regularization scheme that builds on the fact that mass density is more stably imaged. Simulation results demonstrate clear advantages associated with the use of both scattered photon data and energy resolved information in mapping the two material properties of interest. Specifically, comparing images obtained using only conventional attenuation data with those where we employ only Compton scatter photons and images formed from the combination of the two, shows that taking advantage of both types of data for reconstruction provides far more accurate results. 
  \end{abstract}

\begin{IEEEkeywords}
Computed tomography, Compton scattering, limited-view applications, energy-resolved detectors, edge-preserving regularization, inverse problems, iterative
reconstruction
\end{IEEEkeywords}

\IEEEpeerreviewmaketitle

\section{Introduction}
\label{sec:I}
\IEEEPARstart{X}{-ray} CT has been used widely in fields ranging from medical imaging \cite{r1} and non-destructive evaluation \cite{r2} to the investigation of the internal structures of geo-materials \cite{r3} and luggage screening \cite{r4}, the application of specific interest in this paper. Motivated by a desire to construct spatial maps of materials properties (in our case, mass density and photoelectric absorption) in these applications, dual- and multi-energy CT acquisition systems \cite{r5} have drawn much attention in recent years due to their ability to provide high quality images and enhanced material characterization. Specifically, the results in e.g., \cite{r6}, \cite{r7}, \cite{r8} suggest that energy resolving systems perform more robustly for material characterization. For example, in the context of medical imaging, a comparative evaluation performed by \cite{r9} between spectral CT and conventional CT shows that spectral CT is  more reliable in clinical applications in terms of image noise, CT numbers \cite{bushberg2011essential} and quality of reconstruction.\par 
Despite these efforts, simultaneous reconstruction of both photoelectric absorption coefficient and mass density (or the closely related property of Compton scatter attenuation \cite{torikoshi2003electron}) is still challenging due to the lack of sensitivity in the data to variations in the photoelectric absorption coefficient as a function of space \cite{r15}. To address this problem, a number of approaches have been considered in recent years. In \cite{r10} a tensor-based dictionary learning method is introduced for material characterization, taking advantage of high correlation of the attenuation map image between different energy channels. In that work filtered backprojection (FBP) reconstruction is applied to obtain the training dictionary and an alternating iterative optimization approach is used for reconstruction. Another tensor-based iterative algorithm which reconstructs spectral attenuation images is introduced in \cite{r11}. There, a multi-linear image model and tensor-based regularization combined with total variation regularization is proposed to enhance the reconstruction results of low energy channels. In \cite{r12}  the fact that attenuation images are highly correlated in different energy channels has again been used to improve low-dose reconstruction CT. It is assumed that a high quality reference image (RI) of the same object is known. The RI is reconstructed either from a set of normal dose images or reconstructed using energy-integrating projections. To reconstruct attenuation coefficient images in the different channels, a patch-based cost function capturing the correlation between the reference and reconstructed images is introduced and is optimized using the simultaneous algebraic reconstruction technique (SART) \cite{r13}.\par
 
Other approaches to stabilize the photoelectric reconstruction were introduced in \cite{r15}, \cite{r14} focusing on the use of structural regularizers. In \cite{r14} high and low energy attenuation data were collected to estimate Compton and photoelectric attenuation coefficients. In that work an edge-correlation regularization is proposed to aid in the recovery of the  photoelectric coefficient. The same data collection scenario is considered in \cite{r15} to characterize materials in luggage screening application. There a non-local mean (NLM) patch-based regularization scheme was employed to stabilize the recovery of the photoelectric coefficient and an alternating direction method of multipliers (ADMM) method was used to solve the resulting variational problem. In \cite{r16} both photoelectric and Compton attenuation coefficients are recovered from attenuation data for different energy bins by applying a linear mapping function between images of different energy bins to minimize the difference between those images. Also, total variation and the mean of the spectral images are combined to improve the performance of the algorithm. Instead of replacing the conventional integrating detectors with the photon-counting detectors in the hardware domain, a software solution is introduced in \cite{r17} to exploit the information embedded in attenuation data over different energy channels. This method provides the spectral attenuation information by a sparse representation of the reconstructed image at each iteration in a framelet system. \par
The methods cited in the previous two paragraphs focus on cases in which either full view data are provided or, at worse, a limited number of sources (and associated detectors) which fully encircle the object are available for generating data. Reconstruction of photoelectric coefficients in applications with more severely limited-view geometries is more challenging. In many security applications including luggage screening and kVp spectral CT \cite{r18} access to the object from different views are limited while material characterization remains quite critical. In \cite{r18} a maximum-likelihood model employing patch-based regularization is proposed to estimate attenuation coefficients and to exploit the similarity between images from different energy channels. An alternating optimization approach is applied to reconstruct attenuation coefficient images for a set of kVp switching-based sparse spectral CT experiments. In another kVp switching spectral CT application \cite{r19}, attenuation coefficient images are transformed to the Fourier domain and presented in the form of a low-rank Hankel matrix with missing elements. Taking advantage of the high correlation of spectral attenuation images and sparsity in the Fourier domain, the missing elements are recovered by applying SVD-matrix minimization using ADMM. In \cite{r20} an iterative algebraic reconstruction method is proposed for sparse-view CT in medical applications which uses discrete shearlet transformation (DST) for denoising the reconstructed attenuation image at each iteration. Also the effective number of views is increased by interpolating the existing angular views at each iteration.\par 
In this paper, we consider an alternate approach to mapping mass density and photoelectric absorption using both attenuation and Compton scatter data. 
It has been shown that Compton scatter tomography has several advantages over conventional CT systems in e.g., nondestructive evaluation applications \cite{r21}. Compton tomography also provides a powerful tool for materials characterization \cite{r22}. More specifically, Compton scattering is sensitive to structural and density variation within the object \cite{pfeiffer2008hard} by providing a strong contrast mechanism compared to total attenuation \cite{r32}. Most of the Compton scattering tomography reconstruction methods can be divided into analytical and numerical approaches.  A comprehensive review of the analytical solutions is provided in \cite{r23}. The ideas introduced in \cite{r24} are the basis of most of the research in the analytical domain. It has been  shown in \cite{r24} that the scattered beams collected by detectors located on a circular arc connecting the source to the detector, called the `isogonic line', allows for a closed form reconstruction algorithm not unlike conventional filtered backprojection. In a related study, a Radon-transform-like model for a rotating single source/single detector system is introduced in \cite{r25} and provides a closed form solution for recovering the electron density on the arcs passing through the source and detector for each point inside the object. 
Further developments in \cite{r26} show that a Chebyshev integral transform is also applicable to the arcs passing through each point inside the object, which confirms the results provided by \cite{r25}. The same idea has been employed in \cite{r27} for luggage screening applications. There it was shown that a combination of the proposed method and conventional attenuation tomography can produce a map of atomic number. However the approach is not robust to noise, necessitating the use of an ad-hoc pre-processing step of smoothing of the data. In \cite{r28} an analytic approach is proposed for reconstruction of electron densities of tissues for medical applications. \par
Although the analytical methods provide efficient, closed form solutions, they can only be applied to very specific data acquisition geometries. Alternatively, numerical methods such as those considered here provide the flexibility to robustly process data for more general systems. 
In terms of the numerical methods for Compton scatter tomography, most of the work has focused on recovering either the electron density or the total attenuation. A generalized Compton scattering transform that falls in the first category was proposed in \cite{r29} to reconstruct the attenuation map of the object of interest. The energy dependency of the attenuation coefficient at the scattering point was not considered there. In \cite{r30}, the energy dependency of the attenuation is taken into account by approximating the attenuation as a linear function of energy. The algorithm tried to recover the total attenuation coefficient with an iterative minimization method and performed robustly in the presence of noise. The linear approximation to the attenuation holds in the cases that the range of energy change is small. One of the few studies seeking to recover the electron density combines three different interactions, namely fluorescence, Compton scatter and absorption \cite{r31} to directly estimate the unknown fluorescence attenuation map using Compton scattering measurements. Another approach in X-ray Compton tomography assumes the attenuation coefficient is known a priori, from a traditional CT scan, resulting in a linear mapping from density to observations \cite{r32}. Most of the research performed in Compton scattering tomography has focused on gathering the scatter data on energy integrating detectors. In recent years new detectors with good energy resolution have been developed, and a valuable contribution of this paper is exploring how those capabilities can be used in the context of Compton scatter tomography. \par
In most of the work performed in the context of energy-resolved systems, only conventional attenuation data has been considered while the majority of Compton scatter-based imaging has focused on the recovery of attenuation coefficients. In this paper we propose an inversion scheme considering both Compton scattering  and conventional attenuation data for applications with energy-resolved detectors and limited-view geometries to reconstruct both density and photoelectric coefficients. We consider a two-dimensional form of the problem in which scattered photons are collected along with conventional attenuation measurements. A cyclic descent approach is used where we alternate between estimating spatial maps of density and photoelectric attenuation. A multi-scale method is developed to provide an initial estimate of density.   
An edge-preserving method developed in \cite{r36} is employed to regularize the recovery of the density. In order to stabilize photoelectric reconstruction we apply a NLM batch-based regularization \cite{r15}. To evaluate the performance of the proposed method we produce several synthetic phantoms. The simulation results suggests that including Compton scattering tomography as another source of information along with conventional attenuation data can significantly enhance materials characterization especially in challenging applications with limited view geometries.  \par
 The remainder of this paper is organized as follows. In Section \ref{sec:II}, we define a limited-view system and introduce the models we use for both energy resolved attenuation and Compton scatter data. In Section \ref{sec:III} we describe the optimization problem and the iterative reconstruction method for density and photoelectric coefficients. Also, the gradient-based and edge-preserving regularization for density reconstruction and NLM patch-based regularization for photoelectric reconstruction is described. In Section \ref{sec:IV} simulation results are presented and discussed. Section \ref{sec:V} provides concluding remarks and future directions. Finally in the Appendix, we elaborate on the derivative of the cost function and regularization terms required in Levenberg-Marquardt optimization method. 
 \section{Problem Formulation} 
 \label{sec:II}
\begin{figure}[!ht]
  \centering
  \includegraphics[width=3.5in]{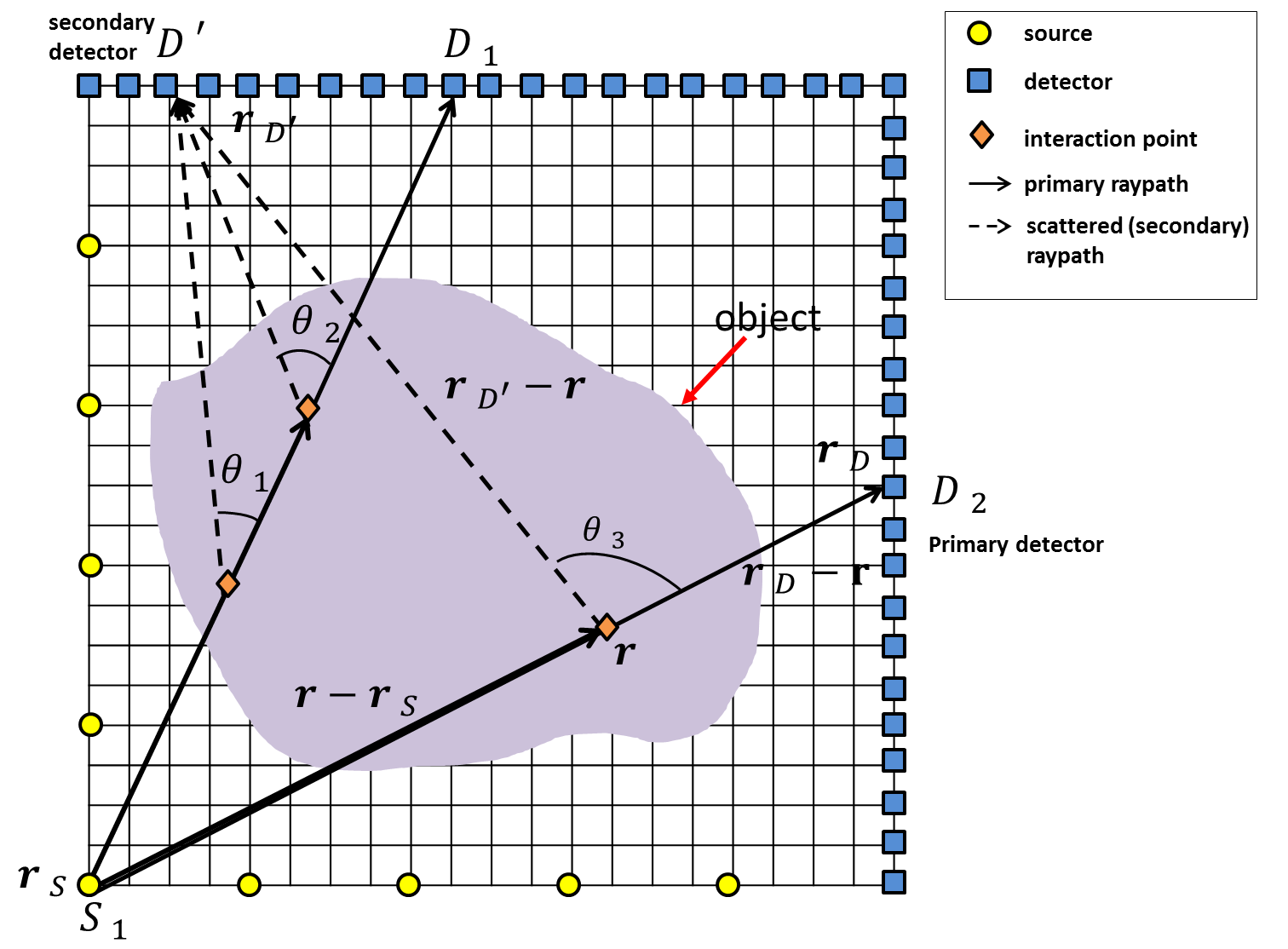}
  \caption{Setup of the sources and detectors. A ray from source $S_1$ to primary detector $D_2$ is scattered with angle $\theta_3$ at the interaction point $r$ and is absorbed by secondary detector $D'$.}
  \label{fig:1}
\end{figure}

 As illustrated in Fig. \ref{fig:1}, here we consider the recovery of mass density and photoelectric absorption in a plane (i.e., a two dimensional problem) based on attenuation and Compton scatter data.  While a 2D physical model for attenuation is commonly employed, Compton scattering is an inherently three dimensional process in that even for strictly ``planar" objects, photons will be scattered into the third dimension. As discussed below, the model we develop accounts for this process and provides an accurate approach for modeling the 2D problem. \par
Shown in Fig. \ref{fig:1} are two types of raypaths and detectors which will be used repeatedly in the rest of the paper. We assume that X-ray sources are collimated to produce pencil beams that illuminate the region of interest, and that these sources are rotated step-wise in angle to produce a set of X-ray beams. Two such beams are shown in Fig. \ref{fig:1}. We refer to an X-ray pencil beam produced by a source traveling through the object on a straight line to a detector as a \textit{primary} raypath and the associated detector a \textit{primary} detector. The number of sources and detectors, $N_S$ and $N_D$, determine $N_{SD}= N_S \times N_D$, the total number of primary raypaths over which attenuation data will be collected. We note that the attenuation data collected along these primary raypaths constitute a typical data set for attenuation-based X-ray imaging methods. \par
\begin{figure}[!ht]
 \centering
  \includegraphics[width=2.5in]{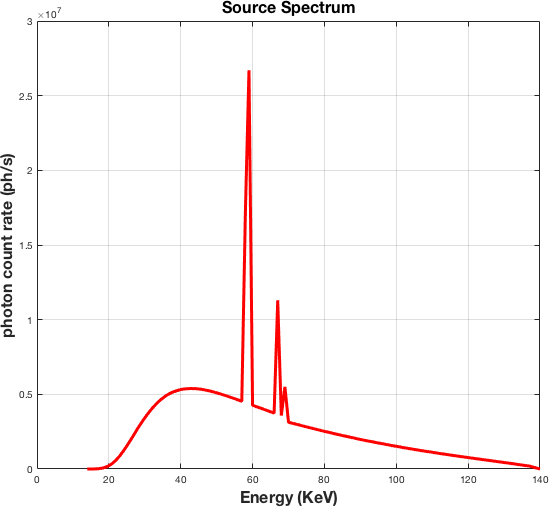}
 \caption{X-ray energy spectrum of a pencil beam source. The $y$ axis shows the number of photons and the $x$ axis shows the energy levels varying from $0$ to $140 \, KeV$ within 1 $keV$ energy bins.}
  \label{fig:2} 
\end{figure}

The Compton scatter data we use for reconstruction as generated by scattering of pencil beam photons at  ``interaction'' points along the primary raypath passing through the object.  At each interaction point along the primary raypath the beam scattered along the \textit{secondary} raypath is observed by a \textit{secondary} detector. As photons travel from the scattering point to the secondary detector, they are further attenuated. For each primary raypath $i=1,\dotsc,N_{SD}$ the total attenuated beam intensity caused by scattering is calculated for each secondary detector $D_{j'}, j'\in \{1,2,\dotsc,N_D \}\setminus\{i\}$,  as shown in Fig. \ref{fig:1}.
 Thus, Fig. \ref{fig:1} illustrates the secondary raypaths connecting two interaction points along the $S_1-D_1$ primary raypath to a detector at $D'$.  At later beam positions, the same detector will measure (as a separate observation) scattering from interaction points along those beams, for example the $S_1-D_2$ path illustrated in Fig. \ref{fig:1}.  
We assume that sources are capable of producing pencil beams which operates over a continuous range of energies. The source energy spectrum $E_{S}$ obtained from \cite{r14} is shown in Fig. \ref{fig:2}. 
We consider detectors of finite energy resolution so that data are retained only in a band of width $\Delta E$ around each $E_m,m=1,2,\dotsc,N_E$ at detectors.
Given this general system setup, we discuss the forward models associated with both absorption and scattering data in the following sections.
  \subsection{Attenuation Tomography Model}
  \label{sec:II.A}
For a given primary raypath the total attenuated beam intensity is calculated at each detector as \cite{r14}
 \begin{equation}
  g(\textbf{r}_{S},\textbf{r}_{D})=\int I(E_S)\left[ \exp\left(-\int\mathrm \mu(\textbf{r}',E_S)\delta_{\textbf{r}_D,\textbf{r}_S}(\textbf{r}') \mathrm{d}\textbf{r}'\right) \right]\,\mathrm{d}E_S
  \label{eq:1}
\end{equation}
where $I(E_S)$ is the intensity of the X-ray source at energy $E_S$, $\delta_{\textbf{r}_D,\textbf{r}_S}(\textbf{r})$ is a Dirac delta function supported along the primary raypath connecting the source position $\textbf{r}_S$ to the detector located at $\textbf{r}_D$, and $\mu(\textbf{r},E_S)$ is the absorption coefficient at energy $E_S$. In the case of energy-discriminating detectors, the energy integral in (\ref{eq:1}) is over the energy bandwidth of a particular energy channel of the detector, while for traditional energy-integrating detectors it is over all energy. As stated earlier the goal of this problem is material characterization which requires in our case recovery of mass density and photoelectric absorption coefficient which are related to $\mu$ according to \cite{alvarez1976energy}
 \begin{equation}
  \mu(\textbf{r},E_S)=N_A \frac {Z(\textbf{r})}{A(\textbf{r})}f_{KN}(E_S)\rho(\textbf{r})+f_p(E_S)p(\textbf{r})
    \label{eq:2}
\end{equation}
where $\rho(\textbf{r})$  is the mass density, $N_A$ is the Avogadro number, $Z(\textbf{r})$ and $A(\textbf{r})$ are the atomic number and atomic weight, $p(\textbf{r})$ is the photoelectric coefficient,
 $f_p (E_S)= E_S^{-3}$ and $f_{KN} (E_S)$, the Klein-Nishina cross section is   
  \begin{equation}
 f_{KN} (E_S)=\frac{1+\gamma}{\gamma^2} \left[\frac{2(1+\gamma)}{(1+2\gamma)}-\frac{1}{\gamma} \ln(1+2\gamma) \right]+\frac{1}{2\gamma}  \ln(1+2\gamma)-\frac{1+3\gamma}{(1+2\gamma)^2} 
 \label{eq:3}
\end{equation} 
where $\gamma=\frac{E_S}{(m_e c^2 )}$. The ratio $\frac{Z(\textbf{r})}{A(\textbf{r})}$ can be approximated to $\frac{1}{2}$ for most of the elements \cite{r31}; therefore (\ref{eq:2}) can be summarized as 
 \begin{equation}
  \mu(\textbf{r},E_S)=\frac {N_A}{2}f_{KN}(E_S)\rho(\textbf{r})+f_p(E_S)p(\textbf{r}).
    \label{eq:4}
\end{equation}
 In the event that detectors are perfectly energy resolving, the polychromatic projection can be replaced by a monochromatic projection so attenuated intensity given in (\ref{eq:1}) can be reduced to a collection of linear systems (one system per energy) relating data to the unknown density and photoelectric absorption coefficient \cite{r11}. For the problem of interest in this paper however, we consider detectors of finite energy resolution. For the imaging method considered in Section \ref{sec:III}, a linear model for attenuation is rather convenient. Toward that end, we consider the following discretized model for the attenuation data which exploits the fact that the energy dependence of the coefficients in (\ref{eq:2}) are well approximated as constant over the ``bins" seen by the detectors even if $I(E_S)$ varies more rapidly. 
 
To discretize the attenuation model, we assume that the object area is discretized on a Cartesian grid with $N_p=N\times N$ elements as shown in Fig. \ref{fig:1}. The system matrix $\textbf{A}$ is then defined where $[\textbf{A}]_{ij}$ represents the length of that segment of primary raypath $i$ passing through pixel $j$ and $[\textbf{A}]_i$ is the $i$-th row of $\textbf{A}$. The size of $\textbf{A}$ is given as $N_{SD} \times N_p$, the product of the number of primary raypaths and number of pixels. 
For each primary raypath $i=1,\dotsc,N_{SD}$ with detector energy bin $E_m,m=1,\dotsc,N_E$ and bandwidth of $\Delta E$, the discrete equivalent to (\ref{eq:1}) is  

\begin{equation}
  g(i,m)=\int_{E_m-\frac{\Delta E}{2}}^{E_m+\frac{\Delta E}{2}} I(E_{S})\left[ \exp \left(-[\textbf{A}]_i\boldsymbol \mu(E_{S})\right) \right] \,\mathrm{d}E_S
 \label{eq:5}
\end{equation}
where $\boldsymbol \mu(E_{S})$ is the lexicographically ordered vector of attenuation coefficients at energy level $E_{S}$. \par
\begin{figure}[!t]
 \centering
\includegraphics[width=3.5in]{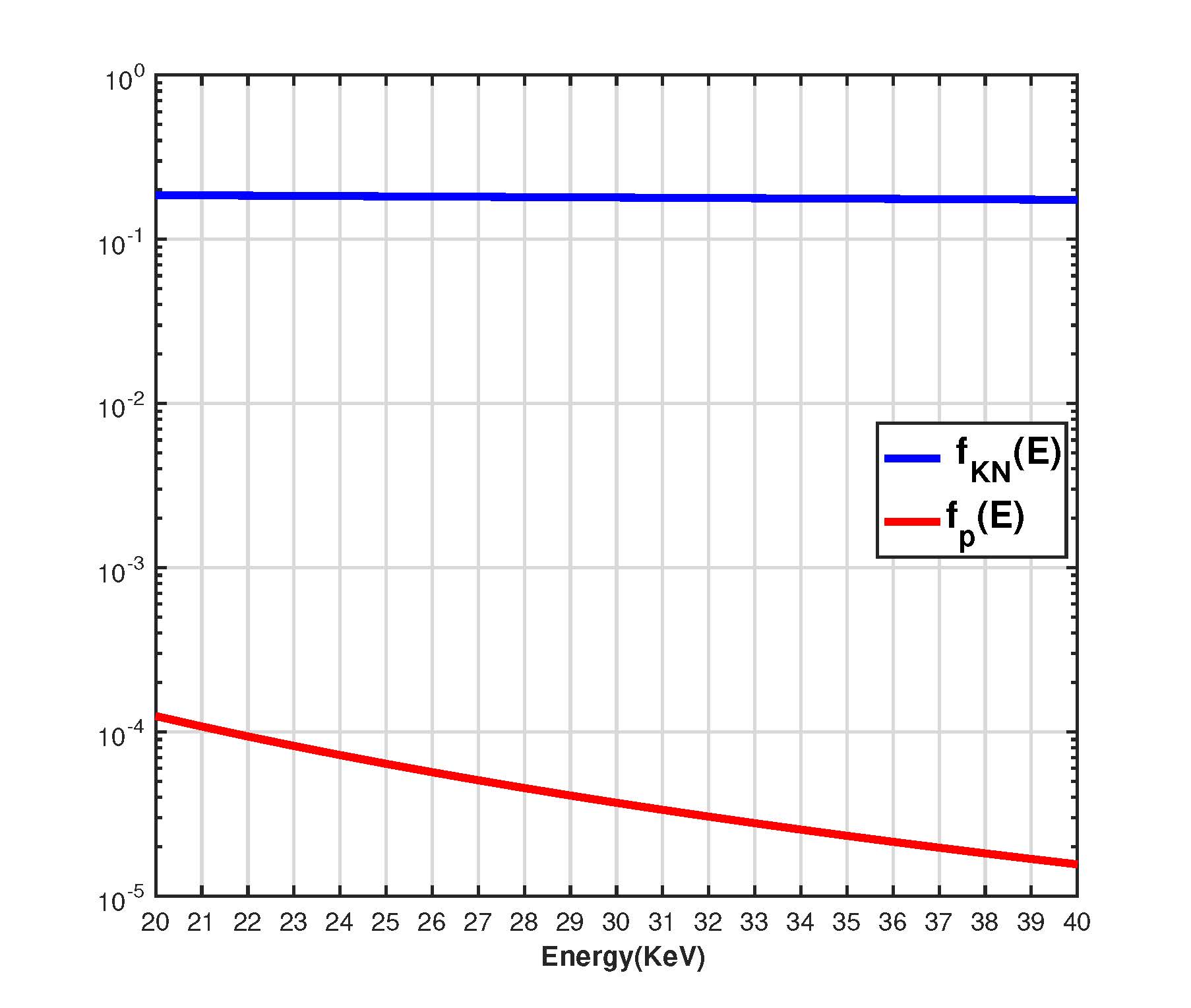}
 \caption{Energy dependent coefficients in mass attenuation. Comparison of Klein-Nishina cross section coefficient $f_{KN} (E_S)$ with $f_{p}(E_S)$. The vertical grid shows the $1 KeV$ bins over which the detectors in this study aggregate photons.}
  \label{fig:3} 
\end{figure}

Referring to (\ref{eq:2}), the terms that depend on energy Klein-Nishina cross section $f_{KN} (E_S)$ and $f_p (E_S)$ are plotted as functions of energy in Fig. \ref{fig:3}. Two characteristics of these graphs are important to us. First, $f_{p}(E_S)$ is much smaller than $f_{KN}(E_S)$ which implies that the data are much less sensitive to photoelectric variations than those of density, a fact we shall exploit in Section III when we discuss the imaging algorithm. Second, both of the functions vary little over the $1KeV$ windows (shown by the vertical lines in Fig. \ref{fig:3}) over which the detectors in this study integrate energy. Thus we replace $\boldsymbol\mu(E_{S})$  with $\boldsymbol\mu(E_{m})$ so that the term $\exp\left(-[\textbf{A}]_i\boldsymbol \mu(E_{m})\right)$ can be factored out of the energy sum. Now, (\ref{eq:5}) simplifies to 
\begin{equation}
  g(i,m)\approx \left[\exp\left(-[\textbf{A}]_i\boldsymbol \mu(E_{m})\right) \right]\int_{E_m-\frac{\Delta E}{2}}^{E_m+\frac{\Delta E}{2}} I(E_{S} )\,\mathrm{d}E_S
  \label{eq:6}
\end{equation}
from which we obtain the following model which is linear in the unknowns of interest: 

\begin{equation}
g_{A}(i,m)=-\log\left(\frac{\textbf{g}(i,m)}{\tilde{I}_m}\right)=[\textbf{A}]_i\boldsymbol \mu(E_{m})
\label{eq:7}
\end{equation}
where $\tilde{I}_m=\int_{E_m-\frac{\Delta E}{2}}^{E_m+\frac{\Delta E}{2}} I(E_{S} )\,\mathrm{d} E_S$. After substituting $\boldsymbol \mu(E_{m})$ given by (\ref{eq:2}), a set of  linear equations with respect to density and photoelectric coefficients is obtained as
\begin{equation}
\textbf{g}_{A}=\textbf{K}_{A,\rho}\boldsymbol \rho+ \textbf{K}_{A,p}\textbf{p}
\label{eq:8}
\end{equation}
where $\textbf{K}_{A,\rho}$ is the discretized attenuation-density system matrix obtained from the terms $\frac {N_A}{2}f_{KN}(E_m)[\textbf{A}]_i$, $\textbf{K}_{A,p}$ is the discretized attenuation-photoelectric system matrix defined by $f_p (E_m )[\textbf{A}]_i$, for $i=1,\dotsc,N_{SD}$ and $m=1,\dotsc,N_E$, and $\boldsymbol\rho$ and $\textbf{p}$ are lexicographically ordered vectors of density and photoelectric images respectively. The vector $\textbf{g}_{A}$ consists of all of the observed attenuation data as a function of source location, primary detector location and energy. The number of elements in $\textbf{g}_{A}$ is equal to $N_{A} = N_{SD} \times  N_E$, the product of the number of primary raypaths $N_{SD}$ and energy bins $N_E$.
\subsection{Scattering Tomography Model}
\label{sec:II.B}
Again referring to Fig. \ref{fig:1}. in this paper, the Compton scattering model captures three physical processes \cite{evans1955atomic}
 \begin{enumerate}
\item  X-ray attenuation from the source to the interaction point along the line connecting $\textbf{r}_S$ and $\textbf{r}$.
\item Compton scattering at the interaction point $\textbf{r}$.
\item Attenuation from the interaction point to the secondary detector $D'$ along the line connecting $\textbf{r}$ and $\textbf{r}_{D'}$.
 \end{enumerate}
Mathematically, we capture these three processes using the following model \cite{r31}
 \begin{equation}
  g_{C}(\textbf{r}_{D'},E')=\int\mathrm I(E_S)\left[\int h(\textbf{r}_{D'},\textbf{r},E')S(\textbf{r},\theta,E_S)f(\textbf{r},\textbf{r}_S,E_S)\delta_{\textbf{r}_D,\textbf{r}_S}(\textbf{r})\rho(\textbf{r}) \mathrm{d}\textbf{r}\right]\,\mathrm{d}E_S
  \label{eq:9}
\end{equation}
  where 
 \begin{itemize}
\item$f(\textbf{r},\textbf{r}_S,E_S)$ is the attenuation of the beam intensity at energy $E_S$  along the line connecting $\textbf{r}_S$ and $\textbf{r}$.
\item$h(\textbf{r}_{D'},\textbf{r},E')$ is the attenuation of the scattered beam at energy $E'$.  We describe below the relationship between $E'$, the energy of the photon emerging from the scattering event, and $E_S$, the initial energy of the photon.
\item$\rho(\textbf{r})$ is the mass density at the interaction point.
\item$S(\textbf{r},\theta,E_S)$ is the scattering factor. We discuss below the relationship between the incident energy of the photon, $E_S$, and the scattering angle, $\theta$.
\end{itemize} 

\begin{figure}[!ht]
  \centering
  \includegraphics[width=3.5in]{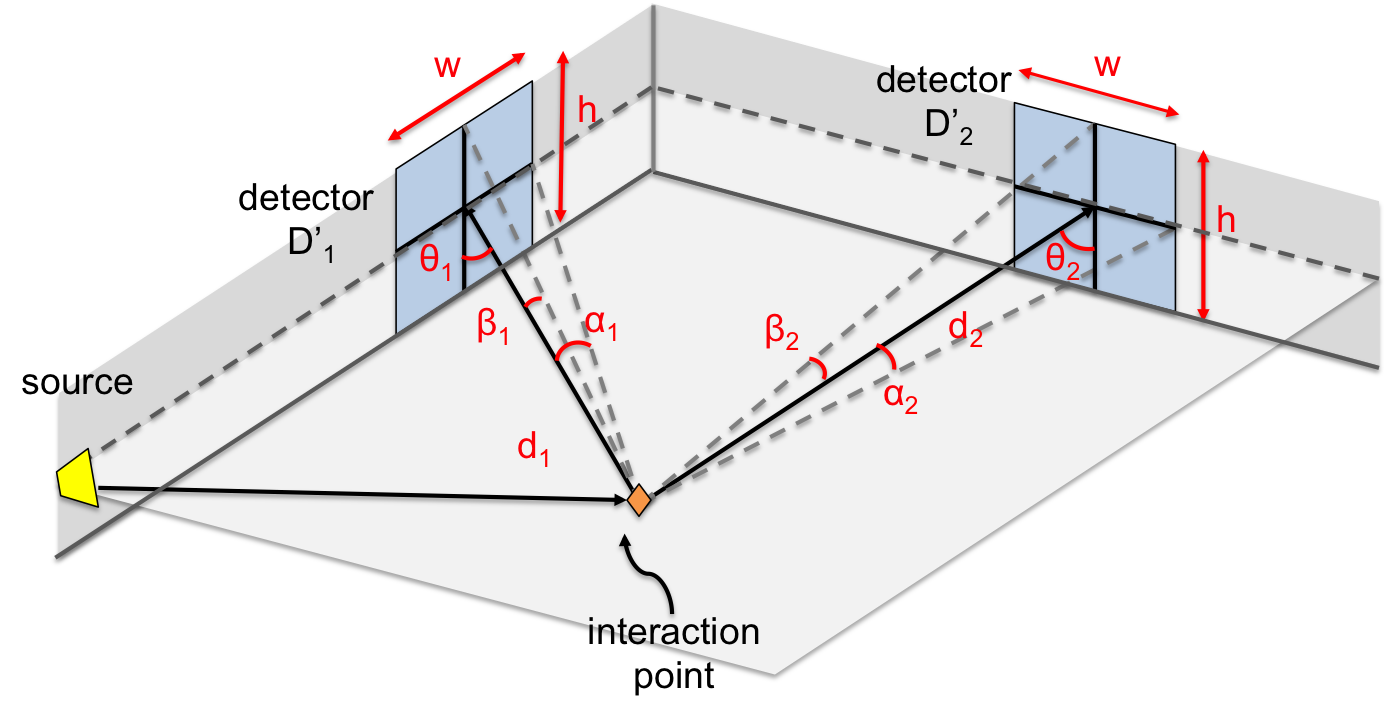}
  \caption{Two rectangular detectors placed in two different locations centered on the plane where we assume all scattering is taking place.
 A ray emitted by the source is scattered to different secondary detectors $D'_1$ and $D'_2$. The height $h$ and width $w$ of the detector, the distance $d$ from the interaction point to detector and relative angles $\alpha$, $\beta$ and $\theta$ determine the solid angle for each interaction point-detector pair. }
  \label{fig:4}
\end{figure}

As in Section \ref{sec:II.A}, attenuation of the beam intensity along the line connecting $\textbf{r}_S$ and $\textbf{r}$ is a function of absorption coefficient  $\mu(\textbf{r},E_S )$ and takes the form
 \begin{equation}
 f(\textbf{r},\textbf{r}_S,E_S )=\exp\left(-\int\mathrm \mu(\textbf{r}',E_S)\delta_{\textbf{r},\textbf{r}_S}(\textbf{r}') \mathrm{d}\textbf{r}'\right)
 \label{eq:10}
\end{equation}
The attenuation of the beam from the interaction point to the secondary detector is much the same except for the fact that the Compton scatter process is inherently three dimensional; i.e., photons are typically removed from the plane of scattering \cite{hartemann2001three}. To capture this effect, we must be a bit more careful with our modeling of the detectors. Specifically, as shown in Fig. \ref{fig:4}, we ascribe to each detector a height and width. Only those photons scattered within the solid angle subtended by the detector are in fact observed \cite{r35}. With this, the attenuation of the beam along the line connected $\textbf{r}$ and $\textbf{r}_D'$  is \cite{r31}:
 
 \begin{equation}
 h(\textbf{r}_{D'},\textbf{r},E' )= \Omega_{D'}(\textbf{r}) \exp\left(-\int\mathrm \mu(\textbf{r}',E')\delta_{\textbf{r}_{D',\textbf{r}}}(\textbf{r}') \mathrm{d}\textbf{r}'\right)
 \label{eq:11}
\end{equation}
where $\Omega_D'(\textbf{r})$ is the solid angle subtended by detector $D'$. In the case of rectangular detectors, $\Omega_D'(\textbf{r})$ is given by \cite{r35}
  \begin{equation}
 \Omega_{D'}(\textbf{r})= 4\arcsin \left(\sin\left(\alpha\right) \times \sin \left(\beta\right) \right)
 \label{eq:12}
\end{equation}
where $\alpha=\arctan\left(\frac{w}{2d}\right)$, $\beta=\arctan\left( \frac{h\cos\theta}{2d}\right)$ and $\theta$ are angles defined in Fig. \ref{fig:4} for two different secondary detectors $D'_1$ and $D'_2$, $h$ and $w$ are height and width of a rectangular detector respectively and  $d$ is the distance from the interaction point to the center of detector area. 

To describe the scattering factor requires a bit of background regarding Compton scattering, an inelastic interaction in which an incident X-ray photon transfers a portion of its energy to a bound electron of the material being probed and emerges at an angle $\theta$ with respect to the initial direction.  As described in \cite{r39} the relationship among the energy of the incident photon, $E_S$, the energy of the scattered photon, $E'$, and $\theta$ is  
  \begin{equation}
  E'=\frac{E_S}{1+\gamma \left(1-\cos(\theta(\textbf{r},\textbf{r}_D,\textbf{r}_{D'}))\right)}
  \label{eq:13}
\end{equation}
where $E_S$ is the incident energy, and referring to Fig. \ref{fig:1}, $\theta(\textbf{r},\textbf{r}_D,\textbf{r}_{D'})$ the scattering angle which can be calculated based on the position of sources and detectors via
  \begin{equation}
  \theta(\textbf{r},\textbf{r}_D,\textbf{r}_{D'})=\cos^{-1}\left(\frac{\textbf{r}-\textbf{r}_D}{| \textbf{r}-\textbf{r}_D |} \cdot  \frac{\textbf{r}-\textbf{r}_{D'}}{| \textbf{r}-\textbf{r}_{D'} |}\right).
  \label{eq:14}
\end{equation}
In (\ref{eq:14}), $\textbf{r}-\textbf{r}_D$  is the vector from the interaction point $\textbf{r}$ to the detector located at $\textbf{r}_D$ and similarly for $\textbf{r}-\textbf{r}_{D'}$.

The scattering factor, $S(\textbf{r},\theta,E)$, in (\ref{eq:9}) is \cite{evans1955atomic}
\begin{equation}
	S(\textbf{r},\theta,E_S)= \rho_e\frac{\mathrm d \sigma_{KN}(E_S,\theta)}{\mathrm d\Omega}   
\label{eq:15}
\end{equation}
where $\rho_e$ is the electron density and $\frac{\mathrm d\sigma_{KN}(E_S,\theta)}{\mathrm d\Omega}$ is the differential Klein-Nishina cross section which gives the fraction of the X-ray energy scattered at angle $\theta$ as 
\begin{equation}
         \frac{\mathrm d\sigma_{KN}(E_S,\theta)}{\mathrm d\Omega}=\frac{r_e^2}{2\left[1+\gamma(1-\cos \theta) \right]^2 } \left[(1+ \cos^2 \theta)+\frac{\gamma^2 (1-\cos \theta)^2}{1+\gamma(1-\cos \theta)}\right] 	
\label{eq:16}
\end{equation}
where $r_e$ is the electron radius.

\begin{figure}[!ht]
  \centering
  \includegraphics[width=3.5in]{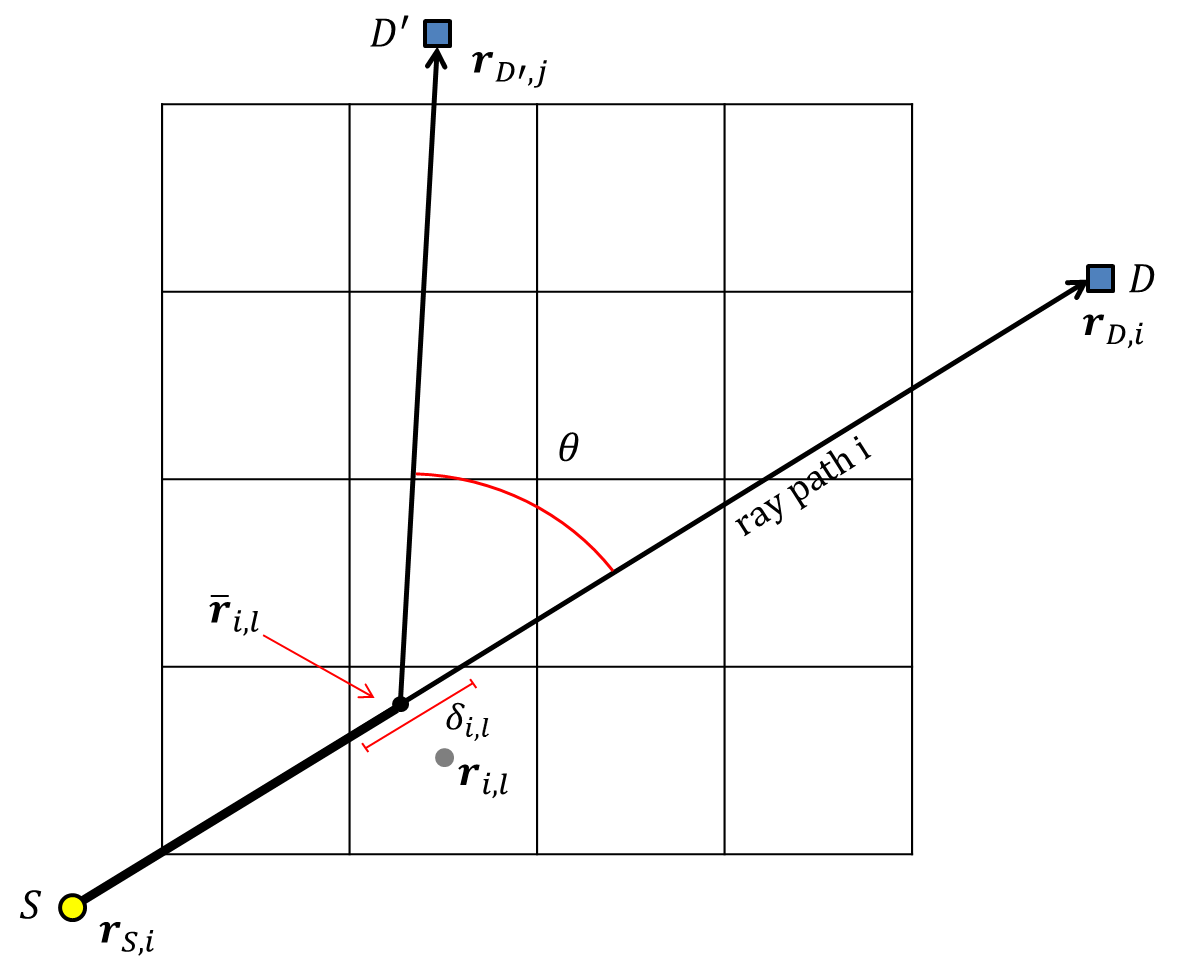}
  \caption{Along the primary raypath $i$ which is from source $S$ to detector $D$ , $\textbf{r}_{i,l}$ is the midpoint of the $l$-th pixel, $\bar{\textbf{r}}_{i,l}$ is the midpoint of the line segment along the primary raypath, and $\delta_{i,l}$ is the length of the line segment.}
  \label{fig:5}
\end{figure}

To discretize (\ref{eq:9}), we approximate the integral over energy using a Riemann sum and employ the same grid used in the case of attenuation data to approximate all spatial integrals to arrive at   
 \begin{equation}
\textbf{g}_{C}(i,j,E'_k)=\sum_{k}{} I(E_{S_k} )\Delta E_S \left[\sum_{l}{}h(\textbf{r}_{D',j},\bar{\textbf{r}}_{i,l},E'_k)S(\bar{\textbf{r}}_{i,l},\theta_{i,j,l})f(\bar{\textbf{r}}_{i,l},\textbf{r}_{S,i},E_{S_k})\delta_{i,l}\rho(\textbf{r}_{j,l}) \right]
\label{eq:17}
\end{equation}
where $\textbf{r}_{D',j}$ is the location of $j$-th secondary detector $D'$, $\textbf{r}_{i,l}$ is the midpoint of the $l$-th pixel on the primary raypath $i$, $\bar{\textbf{r}}_{i,l}$ is the midpoint of the line segment along the primary raypath through this pixel, and $\delta_{i,l}$ is the length of the line segment along primary raypath $i$ crossing this pixel as illustrated in Fig. \ref{fig:5}. 

Because one of the goals in terms of the imaging is the recovery of the mass density, and different rays cross the same pixel in many ways, we assume that the mass density is constant within each pixel and we make the distinction between $\textbf{r}_{i,l}$  and  $\bar{\textbf{r}}_{i,l}$  so that only one unknown will be associated with each pixel but we still provide the correct geometry (specifically, the correct scattering angles) in the specification of the forward model. In this discrete model, attenuation due to absorption between two points is
 \begin{equation}
 f(\textbf{r}_{2},\textbf{r}_{1},E_{S_k})=\exp\left(-\textbf{a}_{\textbf{r}_2,\textbf{r}_1}^T \boldsymbol\mu(E_{S_k})\right)
\label{eq:18}
\end{equation}
where $\textbf{a}_{\textbf{r}_2,\textbf{r}_1}^T$  is a row vector of length $N_p$ whose entries correspond to the length of the line segments crossing each pixel on the path from point $\textbf{r}_1$ to $\textbf{r}_2$
 and  $\boldsymbol\mu(E_{S_k})$ is the lexicographically ordered vector of attenuation coefficients at energy $E_{S_k}$. Finally, the discrete form of the scattering coefficient $S$ is related to the scattering angle and the initial energy such that
 \begin{equation}
 S(\bar{\textbf{r}}_{i,l},\theta_{i,j,l})= \frac{1}{2} N_A  \frac{\mathrm d\sigma_{KN} (E_{S_k},\theta_{i,j,l})}{\mathrm d\Omega }	
 \label{eq:19}
 \end{equation}
where $\theta_{i,j,l}\equiv \theta(\bar{\textbf{r}}_{i,l},\textbf{r}_{D,i},\textbf{r}_{D',j} )$ can be computed using (\ref{eq:14}).
 \par
The inelastic nature of Compton interactions imply that even monochromatic sources will give rise to observed scatter across a band of energies thereby significantly complicating the ``bookkeeping'' associated with this model.  
Because the scattering angle is a function of the energy after the Compton event, the finite bandwidth of our detectors requires that we introduce a window factor in the definition of the scattering coefficient defined in (\ref{eq:19}) as follows
 \begin{equation}
\label{eq:20}	
S(\bar{\textbf{r}}_{i,l},\theta_{i,j,l},E_m)= \frac{1}{2} N_A \frac{\mathrm d \sigma_{KN}(E_{S_k},\theta_{i,j,l})}{\mathrm d\Omega} \omega(i,j,k,l,m)   	
\end{equation}
where, with $E'_k$ defined by (\ref{eq:13}),  
 \begin{equation}
\label{eq:21}	
	 \omega(i,j,k,l,m)=
\begin{cases}
1 & \quad  \theta_{i,j,l} \, \text{such that} \  E'_k \in\left[E_m-\frac{\Delta E}{2},E_m+\frac{\Delta E}{2} \right]\\
 0  & \quad \text{else}\\
\end{cases} 
\end{equation}

Using standard linear algebra, (\ref{eq:17}) can be formulated as a set of equations non-linear in the photoelectric coefficient and quasi-linear in density resulting in a measurement model taking the form
\begin{equation}
\textbf{g}_{C}=\textbf{K}_{C}(\boldsymbol\rho,\textbf{p})\boldsymbol\rho
\label{eq:22}
\end{equation}
where 
$\textbf{K}_{C}(\boldsymbol\rho,\textbf{p})$ is the discretized scattering system matrix obtained from the  terms $h(\textbf{r}_{D',j},\bar{\textbf{r}}_{i,l},E'_k)$, $S(\bar{\textbf{r}}_{i,l},\theta_{i,j,l}))$ and $f(\bar{\textbf{r}}_{i,l},\textbf{r}_{S,i},E_{S_k})$ in (\ref{eq:17}). The vector $\textbf{g}_{C}$ is comprised of all of the observed scattered data as a function of source location, secondary detector location, and energy.  $\textbf{g}_C$ is of size $N_{ST} \times N_E,$ where $N_{ST} = N_S\times N_D\times(N_{D}-1)$ is the number of secondary raypaths, computed from the number of sources $N_S$ and the number of detectors $N_D$, and $N_E$ is the number of detector energy bins.


\subsection{Measurement Noise}
\label{sec:II.C}
While in principle a Poisson model is appropriate for describing both the attenuation and scattered data \cite{sonka2000handbook}, we seek to focus initially on what can be learned from this new class of data in severely limited view geometries. Thus we assume here that the only uncertainty in the data arises from typical additive, white Gaussian noise \cite{tang2009performance}, \cite{siltanen2003statistical}. We leave it to future efforts to extend the ideas developed in this paper to the more complex, but very relevant and interesting, Poisson case. More specifically, the attenuation model after adding noise is defined by  
\begin{equation}
\textbf{g}_A = \textbf{K}_{A,\rho}\boldsymbol\rho + \textbf{K}_{A,p} \textbf{p} + \textbf{w}_A
\label{eq:23}
\end{equation}
where $\textbf{w}_A$ is a white Gaussian noise with zero mean and  variance $\sigma^2_A$. Similarly, the Compton scattering model is given by 
\begin{equation}
\textbf{g}_C = \textbf{K}_C(\boldsymbol\rho,\textbf{p})\boldsymbol\rho + \textbf{w}_C 
\label{eq:24}
\end{equation}
where $\textbf{w}_C$ is a white Gaussian noise with zero mean and variance $\sigma^2_C$ .

 \section{Imaging Approach}
 \label{sec:III}
We propose the following variational problem as the basis for the recovery of density and the photoelectric attenuation coefficient:
\begin{equation}
	(\hat{\boldsymbol\rho},\hat{\textbf{p}} )=\argmin\limits_{\boldsymbol\rho,\textbf{p}} w_1\| \textbf{g}_C-\textbf{K}_C(\boldsymbol\rho,\textbf{p})\boldsymbol\rho \|_{2}^{2}+w_2\|\textbf{g}_A-\textbf{K}_{A,\rho}\boldsymbol\rho-\textbf{K}_{A,p}\textbf{p}\|_{2}^{2}+R_{\rho}(\boldsymbol\rho)+R_p(\textbf{p}|\textbf{I}^{ref})
\label{eq:25}
\end{equation}
where $ \| \textbf{g}_{C}-\textbf{K}_{C}(\boldsymbol\rho,\textbf{p})\boldsymbol\rho \|_{2}^{2}$ measures the mismatch between the scattering data and our prediction of the scattering data for a given $\boldsymbol\rho$ and $\textbf{p}$, and $\|\textbf{g}_{A}-\textbf{K}_{A,\rho}\boldsymbol\rho-\textbf{K}_{A,p}\textbf{p}\|_{2}^{2}$ measures the mismatch between the attenuation data and predicted data. The regularization terms $R_{\rho}(\boldsymbol\rho)$ and $R_p(\textbf{p}|\textbf{I}_{ref})$ for density and photoelectric respectively stabilize the reconstruction by imposing prior information such as smoothness, and $w_1$ and $w_2$ are weighting factors. 
Following \cite{jain2005score} we set $w_1=\frac{1}{\|\textbf{g}_C\|{}_{2}}$ and $w_2=\frac{1}{\|\textbf{g}_A\|{}_{2}}$ to basically normalize the impact of the two data sets in the reconstruction process. \par
We employ a cyclic coordinate descent method \cite{bouman1996unified} for solving the optimization problem given in (\ref{eq:25}). At each iteration, density reconstruction is performed using the estimate of the photoelectric coefficient from the previous iteration. Density reconstruction itself is an iterative procedure detailed below in Section \ref{sec:III.A}. Subsequently, we use the current estimated density image to recover photoelectric coefficient image in another iterative process described in Section \ref{sec:III.B}. 
\subsection{Density Reconstruction} 
\label{sec:III.A}
With $\hat{\textbf{p}}_n$ representing our estimate of the photoelectric coefficient at iteration $n$ of the algorithm, from (\ref{eq:25}), we update the density estimate by solving
\begin{equation}
	\hat{\boldsymbol\rho}_{n+1}=\argmin\limits_{\boldsymbol\rho} w_1\| \textbf{g}_{C}-\textbf{K}_{C}(\boldsymbol\rho,\hat{\textbf{p}}_n)\boldsymbol\rho \|_{2}^{2}+ w_2\| \textbf{g}_{A}-\textbf{K}_{A,\rho}\boldsymbol\rho-\textbf{K}_{A,p}\hat{\textbf{p}}_n\|_{2}^{2}+R_{\rho}(\boldsymbol\rho)
\label{eq:26}
\end{equation}
where the $R_p(.)$ term in (\ref{eq:25}) is not relevant as it does not depend on density. 

In this paper we use an edge-preserving regularization method introduced in \cite{r36}. The approach is based on solving a series of traditional Tikhonov-type smoothness problems where at each iteration, an evolving set of weights is used to decrease the smoothness penalty in regions where edges are suspected. As this method has not appeared in the peer-reviewed literature to date, we provide an overview here. To begin, recall the conventional Tikhonov smoothness-based regularization approach defined as 
\begin{equation}
	R_\rho(\boldsymbol\rho)=\lambda_\rho\|  \textbf{L}\boldsymbol\rho\|_{2}^{2}
\label{eq:27}
\end{equation}
where $\lambda_\rho$ is the regularization parameter which determines the balance between data mismatch and regularization terms,  and $ \textbf{L}$ is a discrete gradient matrix including both vertical and horizontal derivatives computed as 
\begin{equation}
	 \textbf{L}=
\begin{bmatrix}
 \textbf{I} \otimes  \textbf{L}_H \\
 \textbf{L}_V \otimes  \textbf{I}
 \end{bmatrix}
\label{eq:28}
\end{equation}
where $ \textbf{I}$ is an  $N \times N$ identity matrix (assuming we are reconstructing images containing $N_p=N \times N$ pixels), $\otimes$ is the Kronecker tensor product operator and $ \textbf{L}_H =  \textbf{L}_V$  is the $(N-1)\times N$ first difference matrix with $-1$ on the main diagonal and $+1$ on the first upper diagonal.

As noted above, here we employ an approach based on a weighted Tikhonov regularizer for which (\ref{eq:26}) is solved repeatedly. From one iteration to the next the regularization is updated in a manner that de-emphasizes the smoothing for locations in the image where edges are suspected. More specifically, at iteration $l$ the regularization term takes the form

\begin{equation}
R_{\rho,l}(\boldsymbol\rho)=\lambda_{\rho}\|  \textbf{D}^{(l)} \textbf{L}\boldsymbol\rho\|_{2}^{2} \equiv \lambda_{\rho} \|\textbf{M}^{(l)}\boldsymbol\rho\|_2^2
 \label{eq:30}
\end{equation}
where  $\lambda_{\rho}$  is the regularization parameter, $ \textbf{D}^{(l)}=\text{diag}(\textbf{d}^{(l)})$ is a diagonal weighting matrix with elements between zero and one, $ \textbf{M}^{(l)}= \textbf{D}^{(l)} \textbf{L}$, and we call $\textbf{M}^{(l)}\boldsymbol\rho$ the \textit{weighted gradient} of $\boldsymbol\rho$. Those diagonal elements closer to one will enforce smoothness across the associated pixels while the values closer to zero indicate that those pixels belong to an edge and should be preserved. 

To motivate our choice of $\mathbf{d}^{(l)}$, consider a problem like \eqref{eq:26} where now we wish to estimate both $\boldsymbol\rho$ \textit{and} $\mathbf{d}$.  As the elements of $\mathbf{d}$ are non-negative and we expect that most will be close to one and a few closer to zero (since edges are sparse), a reasonable approach for regularizing these quantities would be to employ a entropy-type of functional \cite{fan2010maximum, muniz2000entropy, xu2003minimum}.  In the event that the Boltzman entropy is used for regularizing $\mathbf{d}$ \textit{and} \textit{if} one were to employ a Bregman-type of iteration for estimating $\mathbf{d}$ then (\ref{eq:26}) takes the form 
\begin{equation}
	\label{eq:31}
	\hat{\boldsymbol\rho}^{(l)}_n,\hat{\textbf{d}}^{(l)} = \argmin\limits_{\boldsymbol\rho, \textbf{d}} J_g(\boldsymbol\rho) + \lambda_\rho \|\text{diag}(\textbf{d})\textbf{L}\boldsymbol\rho\|_2^2 + D_{KL}(\textbf{d},\textbf{d}^{(l-1)})
\end{equation}
where $J_g$ is the data fidelty terms in (\ref{eq:26}) and $\textbf{D}_{KL}(\textbf{x},\textbf{y})$ is the generalized Kullback-Leibler divergence defined for non-negative vectors $\textbf{x}$ and $\textbf{y}$ as  $D_{KL}(\textbf{x},\textbf{y}) = \sum_i \textbf{x}_i \log \frac{\textbf{x}_i}{\textbf{y}_i}-(\textbf{x}_i-\textbf{y}_i)$
where e.g.~$\textbf{x}_i$ is the $i$-th element of $\textbf{x}$ \cite{kivinen1997exponentiated}.  (See for example, \cite{burger2016bregman} for the relationship between Boltmzman regularization and a KL-based Bregman problem.)  Using an alternating minimization method for solving (\ref{eq:31}) gives a problem similar in structure to (\ref{eq:26}) for updating the density while a closed form solution for $\textbf{d}$ is easily shown to be
\begin{equation}
	\label{eq:31a}
	\textbf{d}_i^{(l+1)} = \textbf{d}_i^{(l)}\exp\left(-\lambda_\rho\left[\textbf{L}\boldsymbol\rho^{(l)}_n\right]^2_i\right).
\end{equation}
That is, the new estimate for each weight is a scaled version of the old weight where the scale factor is a decreasing function of the strength of the edge at that location. 

Though the ideas in the previous paragraph may be potentially useful in and of themselves for edge-preservation, the exponential dependence yields an approach which is not especially sensitive to edges of varying magnitude. To achieve such sensitivity, we propose the following iteration to replace \eqref{eq:31a}:
\begin{equation}
	\label{eq:31b}
	\textbf{d}_i^{(l+1)} = 
		\textbf{d}_i^{(l)}
			f\left(
					\frac{\left[\textbf{D}^{(l)}\textbf{L}\boldsymbol\rho\right]_i}
				       {\| \left[\textbf{D}^{(l)}\textbf{L}\boldsymbol\rho\right]\|_\infty}
		\right)
\end{equation}
where $f$ is a monotonically decreasing function of its argument whose range is between zero and one.  In this paper we specifically take $f(t) = 1-t^2$.  While we leave the detailed analysis of this method to future work, as partial justification note that in the idealized case where we know the true $\boldsymbol\rho$ at every iteration, the update rule \eqref{eq:31b} gives (for $l \rightarrow \|\textbf{L}\boldsymbol\rho\|_0$, the number of nonzero elements in $\mathbf{L}\boldsymbol{\rho}$),
\begin{equation}
	\textbf{d}_i^{(l)} \rightarrow 
		\begin{cases}
			1 & \left[\textbf{L}\boldsymbol\rho\right]_i=0 \\
			0 & \text{else}
		\end{cases}.
			\label{eq:31c}
\end{equation}
In other words, the vector $\textbf{d}$ acts as an edge detector which, in this case, is zero wherever the gradient is nonzero and one otherwise as is required by an adaptive smoother.  Moreover, as demonstrated and discussed in greater depth in \cite{r36}, the evolution of $\textbf{d}^{(l)}$ in this case puts zeros at locations of large edges in the earlier stages of the iteration while smaller edges are better recovered as $l$ grows.  In a sense then, the approach identifies somewhat coarser structure first and then evolves to recover finer scale details.  \par
We incorporate this approach to regularization into our recovery of $\boldsymbol\rho$ by replacing \eqref{eq:26} with the following:
\begin{equation}
	\hat{\boldsymbol\rho}_n^{(l)} =
		\argmin\limits_{\boldsymbol\rho} w_1\| \textbf{g}_{C}-\textbf{K}_{C}(\boldsymbol\rho,\hat{\textbf{p}}_n)\boldsymbol\rho \|_{2}^{2}+ w_2\| \textbf{g}_{A}-\textbf{K}_{A,\rho}\boldsymbol\rho-\textbf{K}_{A,p}\hat{\textbf{p}}_n\|_{2}^{2}+
								 \lambda_\rho \|\textbf{M}^{(l)}\boldsymbol\rho\|_2^2
\label{eq:31d}
\end{equation} 
which we write in the more convenient form
\begin{equation}
	\label{eq:31e}
	\hat{\boldsymbol\rho}_n^{(l)} = \argmin\limits_{\boldsymbol\rho} \left\| \tilde{\textbf{g}}-\tilde{\textbf{K}}^{(l)}(\boldsymbol\rho)\boldsymbol\rho \right\|_{2}^{2}
\end{equation}
with 
\begin{equation}
	\label{eq:31f}
	\tilde{\textbf{g}} = 
		 \begin{bmatrix} \sqrt{w_1}\,\textbf{g}_{C}\\ \sqrt{w_2 }\,(\textbf{g}_{A}-\textbf{K}_{A,p}\hat{\textbf{p}}_n)\\ 0\end{bmatrix}
		\quad \text{and} \quad
	\tilde{\textbf{K}}^{(l)}(\boldsymbol\rho) =
		\begin{bmatrix} 
		\sqrt{w_1}\,\textbf{K}_{C}(\boldsymbol\rho,\hat{\textbf{p}}_n)\\ 
		\sqrt{w_2}\, \textbf{K}_{A,\rho}\\  
		\sqrt{\lambda_\rho}\, \textbf{M}^{(l)}
		\end{bmatrix}.
		\end{equation}

There remain two issues concerning this approach: how to solve \eqref{eq:31e} and how to terminate the iteration in $l$.  The quasi-linear form of the cost function in \eqref{eq:31e} immediately suggests a fixed point iteration.  Specifically, starting with an initial guess for the density, call it $\tilde{\boldsymbol\rho}$, we  build $\tilde{\textbf{K}}(\tilde{\boldsymbol\rho})$ so that the resulting problem, $\argmin\limits_{\boldsymbol\rho}\|\tilde{\textbf{g}}-\tilde{\textbf{K}}^{(l)}(\tilde{\boldsymbol\rho})\|_2^2$ is a linear least squares problem for $\boldsymbol\rho$.  Due to the size and sparsity of the matrices comprising $\tilde{\textbf{K}}^{(l)}$, the iterative solver LSQR  \cite{r33} is used to find the solution to this problem. That solution is then used to build a new $\tilde{\textbf{K}}^{(l)}$ and the process repeats.  For the problems considered in Section \ref{sec:IV}, this ``inner'' iteration converges rather quickly with the $L_2$ norm of the difference between the density estimates below $10^{-11}$ in roughly $7$ iterations.  \par
The termination of the ``outer'' edge-preserving iteration over $l$ is required due to the monotonically decreasing nature of the diagonal weighting matrix implied by \eqref{eq:31b}.  Indeed, except in  cases where the gradient is \textit{exactly} zero, the weights will, as $n\rightarrow\infty$, go to zero resulting in an unregularized problem. In this paper, we choose to stop when the change in weighted gradient is small or we have exceeded some maximum number of iterations; i.e., when
\begin{equation}
	\|\textbf{M}^{(l+1)}\boldsymbol\rho_n^{(l+1)} - \textbf{M}^{(l)}\boldsymbol\rho_n^{(l)}\|_2^2 < \epsilon_l
	\text{ or } l > l_{\text{max}}.
		\label{eq:31g}
\end{equation}
where $\epsilon_l$ is a small number and $l_{\text{max}}$ is the maximum number of iterations.
For the cases in Section \ref{sec:IV}, typically we see convergence after approximately $10$ iterations. \par
\begin{table}[h!]
\centering
\begin{tabular}{@{} l} \hline 
\textbf{Inputs:} \\
\hspace{35pt} \textbullet \thinspace $ \tilde{\textbf{g}}$ \text{,} $\textbf{K}_{A,p}$ \text{,} $\textbf{p}$ \text{and} $\textbf{L}$ \\ 
\hspace{35pt} \textbullet \thinspace $ w_1$\text{,} $w_2 $\text{,} $\epsilon_{EPI}$ \text{and} $\epsilon_{FPI}$\\ 
\textbf{Initialize:} \\
\hspace{35pt} \textbullet \thinspace $ l=1$ and ${flag}_{EPI}=1$ \% EPI = Edge Preserving Iteration\\ 
\hspace{35pt} \textbullet \thinspace $ \textbf{D}^{(l)}= \textbf{I}$ and $ \textbf{M}^{(0)}= \textbf{I}$ \\ 
\hspace{35pt} \textbullet \thinspace $\boldsymbol\rho_n^{(0)}=$ vector of  $+ \infty$ to force at least one edge-preserving iteration \\ 
\hspace{35pt} \textbullet \thinspace $\textbf{r}_{old}=\boldsymbol\rho_n^{(0)}$ \\ 
\textbf{1:} \thinspace \textbf{While} ${flag}_{EPI}$  true\\
\textbf{2:}\hspace{30pt} Set $ \textbf{M}^{(l)}=\textbf{D}^{(l)}\textbf{L}$\\
\textbf{3:}\hspace{30pt} Set ${flag}_{FPI}=1$ \% FPI = Fixed Point Iteration\\
\textbf{4:}\hspace{30pt} \textbf{While} ${flag}_{FPI}==1$\\
\textbf{5:}\hspace{50pt} Build $\tilde{\textbf{K}}^{(l)}(\textbf{r}_{old})$ according to (\ref{eq:31f})\\
\textbf{6:}\hspace{50pt} Find $\textbf{r}_{new}$ by solving (\ref{eq:31e}) with LSQR \\
\textbf{7:}\hspace{50pt} \textbf{IF}  $ \| \textbf{r}_{new}-\textbf{r}_{old} \|_{2}^{2} <  \epsilon_f$ : \% The inner, fixed point iteration has converged\\
\textbf{8:}\hspace{70pt} Update $\boldsymbol\rho_n^{(l)}= \textbf{r}_{new}$\\ 
\textbf{9:}\hspace{70pt} Set \thinspace $ {flag}_{FPI}=0$ \\
\textbf{10:}\hspace{50pt} \textbf{ELSE} : \\
\textbf{11:}\hspace{70pt} $\textbf{r}_{old} =  \textbf{r}_{new}$\\ 
\textbf{12:}\hspace{30pt} \textbf{end}\\
\textbf{13:}\hspace{30pt} \textbf{IF} $\|\textbf{M}^{(l+1)}\boldsymbol\rho_n^{(l+1)} - \textbf{M}^{(l)}\boldsymbol\rho_n^{(l)}\|_2^2 < \epsilon_l$ or  $l > l_{\text{max}}$ : \% The outer, edge preserving iteration has converged\\
\textbf{14:}\hspace{50pt} Update $\hat{\boldsymbol\rho}_n=\boldsymbol\rho_n^{(l)}$ \\ 
\textbf{15:}\hspace{50pt} Set ${flag}_{EPI}=0$ \\ 
\textbf{16:}\hspace{30pt} \textbf{ELSE} : \\
\textbf{17:}\hspace{50pt} Update $\textbf{d}$ according to (\ref{eq:31b})\\
\textbf{18:}\hspace{50pt} Increase $l$ \\
\textbf{19:}\thinspace \textbf{end} \\
 \hline
\end{tabular}
\caption{Pseudo code for iterative quasi-linear solver}
\label{Table:1}
\end{table}

The pseudo-code in Table \ref{Table:1} summarizes the overall approach for determining $\boldsymbol\rho_n$. 
To begin the process, we require an initial estimate for the density, $\boldsymbol\rho=\boldsymbol\rho_0$ and assume $\textbf{p}=\textbf{0}$ at iteration $n=0$. Starting with an appropriate initial guess for the density is crucial to the success of the approach. We note there are a number of ways this could be accomplished.  For example, attenuation based CT images have been shown to be useful in this regard \cite{r25}. However for the limited view problems that interest most in this effort, reconstruction of the photoelectric and density from attenuation data is known to be a highly ill-posed problem. Thus to improve the convergence rate of the density reconstruction and reduce the overall time complexity, we are motivated to consider an alternate, multi-scale approach which is used only at $n=0$ when we have essentially no prior information regarding the composition of the medium. Specifically, we begin with a coarse spatial representation of the density initialized to a constant value with the same constant used for all experiments in Section \ref{sec:IV}. The method in Table \ref{Table:1} is used to solve the problem at this spatial scale and the estimated density image at this level is ``upscaled" employing nearest neighbor interpolation with the Matlab function `imresize()' and used as an initial guess to build the system matrix at the next finer scale. This multi-scale process continues until we reach the desired, finest scale. 

\subsection{Photoelectric Reconstruction} 
\label{sec:III.B}
Given $\hat{\boldsymbol{\rho}}_n$, the photoelectric subproblem takes the form
\begin{equation}
	\hat{\textbf{p}}_{n+1}=\argmin\limits_{\textbf{p}} w_1\left \| \textbf{g}_{C}-\textbf{K}_{C}(\hat{\boldsymbol\rho}_n,\textbf{p})\hat{\boldsymbol\rho}_n \|_{2}^{2}+w_2\|\textbf{g}_{A}-\textbf{K}_{A,\rho}\hat{\boldsymbol\rho}_n-\textbf{K}_{A,p}\textbf{p} \right\|_{2}^{2}+R_{p}(\textbf{p}|\textbf{I}^{ref})
\label{eq:32}
\end{equation}
where $\hat{\boldsymbol\rho}_n$ is the final estimate of density image at previous iteration as a solution to (\ref{eq:26}) and $R_{p}(\textbf{p}|\textbf{I}^{ref})$ is the photoelectric regularization term. 
In contrast to the density problem, photoelectric recovery is a non-linear least squares optimization problem which we solved using the Levenberg-Marquardt method \cite{r34}. The approach requires the Jacobian matrix of the objective function which is given in Appendix A. \par
It is well known that the recovery of the photoelectric map is a challenging problem \cite{r15,r14} while density is, roughly speaking, far easier to obtain accurately. To stabilize the photoelectric problem, we have used patch-based non-local mean (NLM) regularization method \cite{r15} which benefits from the accuracy with which density can be recovered.   In this approach the photoelectric reconstructed image is conditioned on a reference image $\textbf{I}^{ref}$ which we take as $\hat{\boldsymbol\rho}_{n=1}$, the density estimate obtained after the first iteration of the algorithm. Mathematically, the NLM regularization can be written in the form of quadratic regularization as 
\begin{equation}
	R_{p}(\textbf{p}|\textbf{I}^{ref})=R_{NLM} (\textbf{p}|\hat{\boldsymbol\rho}_{n=1} )= \lambda_p\|(\textbf{I}-\textbf{W})\textbf{p}\|_2^2
\label{eq:33}
\end{equation}
where $\textbf{I}$ is the identity matrix, $\textbf{W}$ is the weight matrix which is calculated based on the reference image \cite{buades2005non},  \cite{darbon2008fast} and $\lambda_p$ is the regularization parameter. By stacking $ \textbf{K}_{C}(\hat{\boldsymbol\rho}_n,\textbf{p})\hat{\boldsymbol\rho}$, $ \textbf{K}_{A,p}\textbf{p}$ and $ (\textbf{I}-\textbf{W})\textbf{p}$ vectors (\ref{eq:32}) takes the form 
\begin{equation}
	\hat{\textbf{p}}_{n+1}=\argmin\limits_{\textbf{p}} \left\| \begin{bmatrix} \sqrt{w_1}\,\textbf{g}_{C}\\ \sqrt{w_2 }\,(\textbf{g}_{A}-\textbf{K}_{A,\rho}\hat{\boldsymbol\rho}_n)\\ 0\end{bmatrix} - \begin{bmatrix} \sqrt{w_1}\, \textbf{K}_{C}(\hat{\boldsymbol\rho}_n,\textbf{p})\hat{\boldsymbol\rho}\\ \sqrt{w_2}\, \textbf{K}_{A,p}\textbf{p} \\  \sqrt{\lambda_p}\,(\textbf{I}-\textbf{W})\textbf{p})\end{bmatrix} \right\|_{2}^{2} \equiv  \argmin\limits_{\textbf{p}} \left\| \tilde{\textbf{q}}-\tilde{\textbf{Q}}(\textbf{p}) \right\|_{2}^{2}.
\label{eq:34}
\end{equation}
The reader is referred to the Appendix for further details of the solution procedure.
\section{Experiment}
\label{sec:IV}
\begin{figure*}[!t]
\centering
\subfloat[Phantom I - Density Image]{\includegraphics[width=2in]{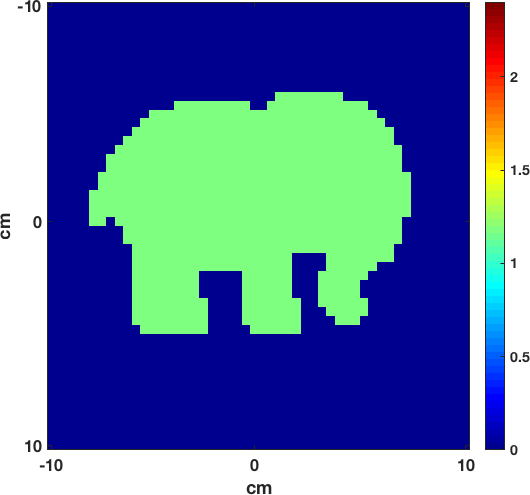}
\label{fig:6a}}
\hfil
\subfloat[Phantom II- Density Image]{\includegraphics[width=2in]{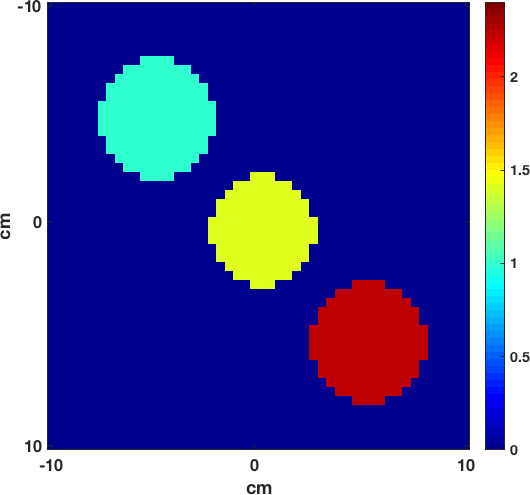}
\label{fig:6b}}
\vfil
\subfloat[Phantom I- Photoelectric Image]{\includegraphics[width=2in]{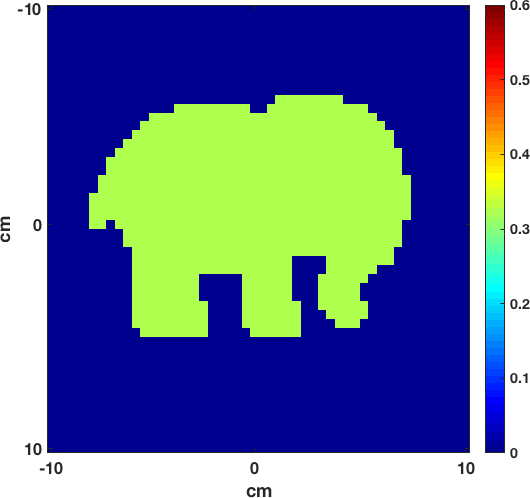}
\label{fig:6c}}
\hfil
\subfloat[Phantom II- Photoelectric Image]{\includegraphics[width=2in]{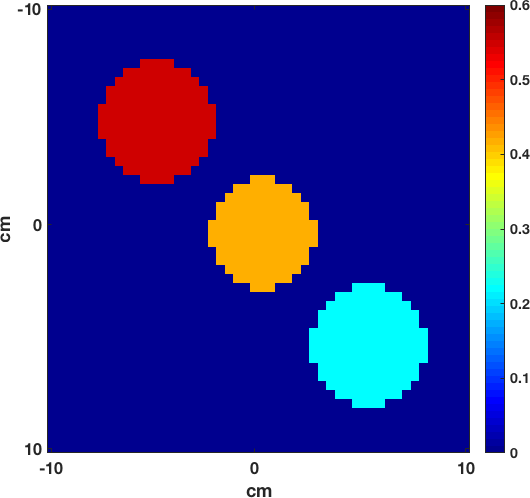}
\label{fig:6d}}

\caption{Simulated phantoms. Density and Photoelectric (at the energy level of $E_0=20\, KeV$) ground truth images of different objects described in Table \ref{Table:2}. (a) and (b) are plotted in the range of $[0,2.4] g/cm^3$ and (c) and (d) are plotted in the range of $[0,0.6] cm^{-1}$.}
 \label{fig:6}
\end{figure*}

To evaluate our proposed method we consider a limited view system of the form provided in Fig. \ref{fig:1}. The area to be imaged is taken to be  $20  \, {cm} \times 20 \, {cm}$.  
Three rotating pencil beam sources each with a spectrum shown in Fig.  \ref{fig:2} are located exactly in the center of the left and  bottom edges and left-bottom corner of the scanning area. 
Forty-one detectors with the width and height of  $0.1 \,cm$ are equally spaced along the top and right edges. All data are generated assuming a uniform grid of $50 \times 50$  pixels covering the $400\, {cm}^2$ region. For the multi-scale processing method described in Section \ref{sec:III.A}, five uniform grids of $10 \times 10$, $20 \times 20$, ..., $50 \times 50$ are employed for the unknown mass density.
We have generated synthetic data for two different phantoms consisting of different materials with moderate to high attenuation properties shown in Fig. \ref{fig:6}. The first phantom in the shape of an elephant \footnote{Tufts' official mascot is Jumbo the elephant: https://www.tufts.edu/about/jumbo} and with the material properties of plexiglass provides an interesting challenge in terms of recovering the intricate geometry of the object due to some rather challenging geometric details (e.g., the space between the legs, the trunk, etc). The second phantom is more complicated with three circular objects consisting of water, Delrin and graphite. The characteristics of the materials used in these phantoms are taken from the XCOM database \cite{xcom} and described in detail in Table \ref{Table:2}. \par

\begin{table}[h!]
\centering
\begin{tabular}{l c r }\hline 
Material & Density  $g/{cm}^3$ & Photoelectric $cm^{-1}$ \\
 \hline
 Delrin & 1.4 & .4134 \\
  Graphite &2.23 & .2177 \\
  Plexiglass & 1.18 & .3263 \\
  Water & 1 & .5439 \\ 
  \hline
\end{tabular}
\caption{density and photoelectric coefficient of objects in simulated phantoms.}
\label{Table:2}
\end{table}

Attenuation data is collected in the range of $20-120 \,KeV$ on the energy resolution of $\Delta E=1 \,KeV$ for density and photoelectric coefficient reconstruction according to (\ref{eq:23}).  Because of the size of the resulting data set ($123$ primary ray paths $\times$ $40$ scatter detectors per raypath $\times$ $100$ energy bins $= 4.92 \times 10^5$ observations), we have chosen to bin the scattered data into $5 \,KeV$ intervals so as to reduce the computational overhead of the processing.  
To consider measurement and discretization noise, a signal-to-noise (SNR) ratio of $50$ dB is assumed for both attenuation and scattering measured data. \par  
All the simulations are performed in MATLAB with the processing architecture of  $8$ core Intel CPU and $50$ gigabytes of memory. The code used in these experiments is not optimized in terms of time and complexity efficiencies. The main computational load belongs to the LSQR solver and calculating forward model and Jacobin matrices, with $ 352 \,sec.$,  $4.6 \,sec.$ and $25.2 \,sec.$ on average per iteration respectively.    
\par In the cyclic descent method described in Section  \ref{sec:III.A},  at each iteration, to reconstruct density the estimates of the photoelectric coefficient and density from previous iteration are required. At the initial iteration, $n=0$, according to (\ref{eq:26}) the estimation of density $\hat{\boldsymbol\rho}_1$ requires photoelectric coefficient $\hat{\textbf{p}}_0$ which we take as $\hat{\textbf{p}}_0=\textbf{0}$. The density is initialized with $\hat{\boldsymbol\rho}_{0} = {.4} \,g/{cm}^3$ for both phantoms.
 For the photoelectric reconstruction at $n=0$, $\hat{\boldsymbol\rho}_1$ is used in (\ref{eq:32}) and the Levenberg-Marquardt method is initialized with $\textbf{p}_0=\textbf{0}$,  where for $n>1$  $\hat{\textbf{p}}_{n-1}$ is used. \par
The regularization parameters $\lambda_\rho$ and $\lambda_p$ discussed in Section  \ref{sec:III.A}  and Section  \ref{sec:III.B} are determined using the discrepancy principle \cite{r37} since the variance of the noise is assumed known. In theory these parameters should be selected by first discretizing the space of both $\lambda_\rho$ and $\lambda_p$, calculating the reconstructions of density and photoelectric for all the points on this two-dimensional discretized space, and then computing the value of the discrepancy function for each of these reconstructions. The optimal parameters and associated reconstructions output by the algorithm would be those associated with the minimum  of the discrepancy function.  
Given the computational burden of the reconstruction process, we choose to employ the following suboptimal method.  At iteration $n=1$ for density reconstruction where $\hat{\textbf{p}}_0=\textbf{0}$, 
each scale of the multi-scale reconstruction process is repeated $25$ times for $25$ logarithmically spaced values of $\lambda_\rho$ between $10^{-4}$ and $10^4$. At each scale, we choose that estimate of density which minimized the discrepancy function 
\begin{equation}
F_{D,\rho}(i,k)= \frac{1}{\tau}\|\textbf{r}_{i,k}\|_2^2 - \sigma^2  \quad i=1, 2, \dots, 25 \quad \text{and} \quad k=1, 2, \dots, 5
\label{eq:40}
\end{equation}
where $\textbf{r}_{i,k}= \tilde{\textbf{g}}-{\tilde{\textbf{K}}_{i,k}}(\boldsymbol\rho_{k})\boldsymbol\rho_{k}$ is the regularized residual of the density reconstruction defined in (\ref{eq:31}), $\tau$ is the number of the elements of the data vector, $i$ is the regularization parameter indicator, $k$ corresponds to the scale level  and $\sigma^2$ is the noise variance. For $n > 1$, we use as $\lambda_\rho$ the value of this parameter associated with the reconstruction selected at the finest scale of the $n=1$ iteration.  
Again at iteration $n=1$, where the density estimation $\hat{\boldsymbol\rho}_1$ is used for reconstruction of photoelectric, an analagous approach is used to determine $\lambda_p$ which is then used for the remainder of the iterations.
Despite the suboptimal nature of this process the quantitative and qualitative measurements of the reconstruction results are highly acceptable given the limited nature of the source/detector geometry. 

The stopping criteria for the overall algorithm is based on the density convergence. Thus, if the current estimation of density satisfies the convergence condition then the reconstruction of photoelectric using the final estimation of density will conclude the cyclic coordinate descent procedure. The stopping criteria is defined as \cite{K.Madsen2004} 
\begin{equation}
	 \|\boldsymbol\rho_{n}-\boldsymbol\rho_{n-1}\|_2^2< \epsilon\left(1+ \|\boldsymbol\rho_{n-1}\|_2^2 \right)
\label{eq:42}
\end{equation}
where $\boldsymbol\rho_{n}$ is the density estimated vector at $n^{th}$ iteration
and $\epsilon$ is a small, positive number defines the accuracy of the final results which is taken $10^{-2}$.The stopping criteria $\epsilon_{FPI}$ for the fixed-point iteration and $\epsilon_{EPI}$ for the edge-preserving procedure defined in Table \ref{Table:1} are taken  as $10^{-11}$ and $3 \times10^{-3}$ respectively, and $l_{max}=100$.\par
To evaluate the performance of the proposed method quantitatively, we have calculated the relative mean square error (RMSE) for each of density and photoelectric images using

\begin{equation}
RMSE= \frac{\|\hat{\textbf{I}} -\textbf{I}_{true}\|_2^2}{\|\textbf{I}_{true}\|_2^2}	
\label{eq:42a}
\end{equation}
where $\hat{\textbf{I}}$ is the reconstruction of either the density or the photoelectric image and $\textbf{I}_{true}$ is the corresponding ground truth image. \par
There are a number of aspects of the reconstruction process we wish to explore with these examples. We first compare the recovered density and photoelectric maps after the first iteration (i.e., $n=1$) of the algorithm. This analysis allows us to explore the utility of the multi-scale method for recovering density. After exploring these issues, we turn to the impact of iterating past $n=1$ and examine improvements seen in our ability to recover both parameters of interest.  Finally, we compare our ability to quantify materials as a function of the data type used in the image formation process.  We note that in all cases, the fusion of scatter data with traditional attenuation greatly improves both the quantitative as well as qualitative characteristics of the processing results. 
 \par We explore the effects of attenuation-only, scattering-only and combination of both datasets in reconstructing density at first iteration by setting $w_1=0$ and $w_2=1$ ,then $w_1=1$ and $w_2=0$ and finally $w_1=\frac{1}{\|\textbf{g}_C\|{}_{2}}$ and $w_2=\frac{1}{\|\textbf{g}_A\|{}_{2}}$ respectively in (\ref{eq:26}). 
 \par
\begin{figure*}[!t]
\centering
\subfloat[Scale 1: RMSE=0.6153]{\includegraphics[width=2in]{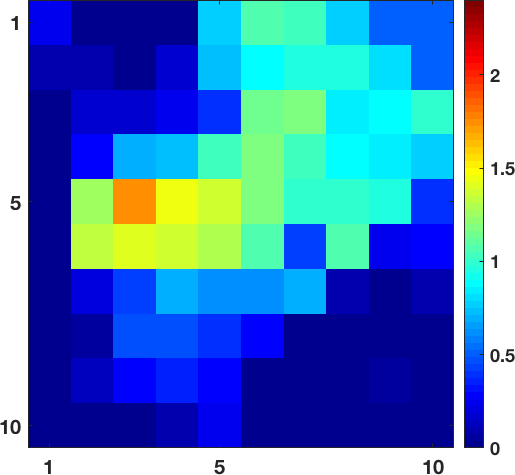}
\label{fig:8a}}
\hfil
\subfloat[Scale 2: RMSE=0.4775]{\includegraphics[width=2in]{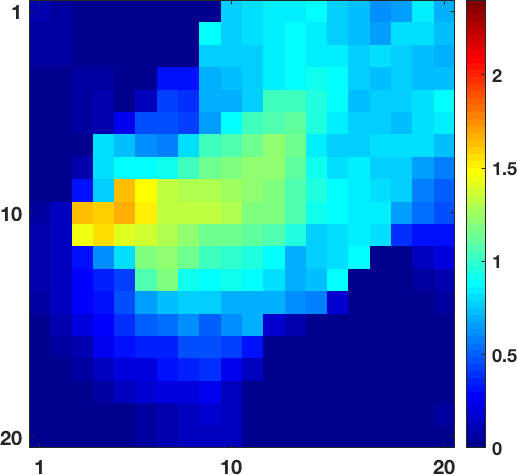}
\label{fig:8b}}
\hfil
\subfloat[Scale 3: RMSE=0.4043]{\includegraphics[width=2in]{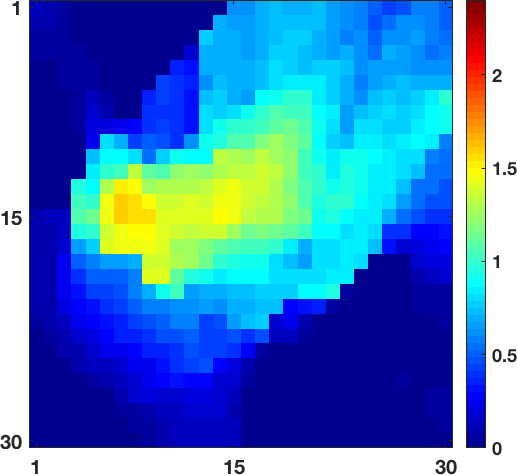}
\label{fig:8c}}
\vfil
\subfloat[Scale 4: RMSE=0.3962]{\includegraphics[width=2in]{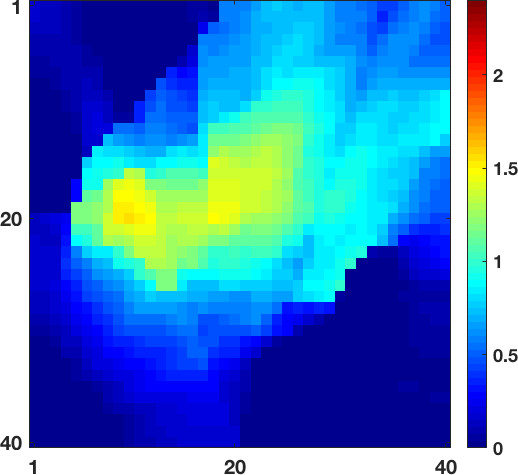}
\label{fig:8d}}
\hfil
\subfloat[Scale 5: RMSE=0.3172]{\includegraphics[width=2in]{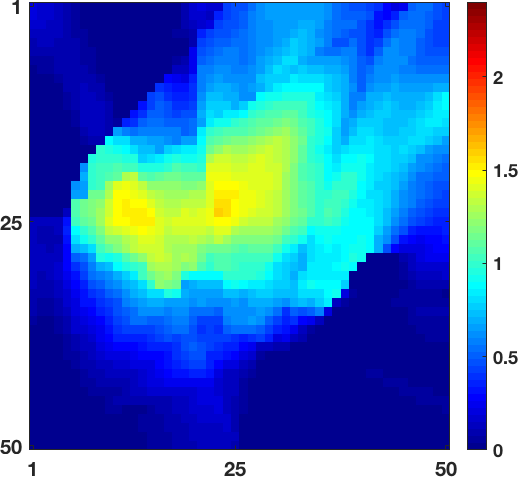}
\label{fig:8e}}

 \caption{ Density reconstruction results obtained using attenuation data alone for Phantom-I at each scale of processing at the first iteration of the algorithm. While the amplitude of the reconstructions are reasonably accurate, geometric structure is less well resolved.  Subplots (a)-(e) show the density reconstruction results for $5$ different grid sizes from $10\times 10$ to $50\times 50$.}
 \label{fig:8}
\end{figure*}

\begin{figure*}[!t]
\centering
\subfloat[Scale 1: RMSE=0.7319]{\includegraphics[width=2in]{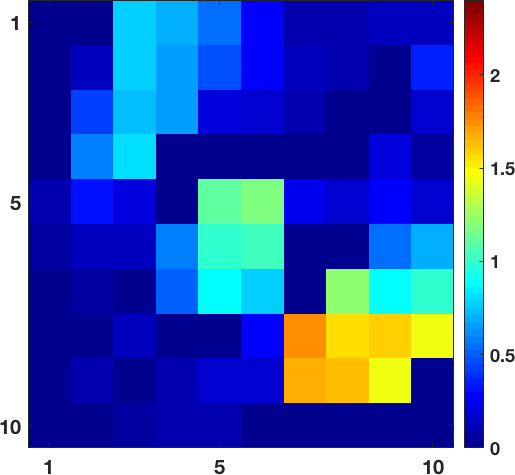}
\label{fig:9a}}
\hfil
\subfloat[Scale 2: RMSE=0.4618]{\includegraphics[width=2in]{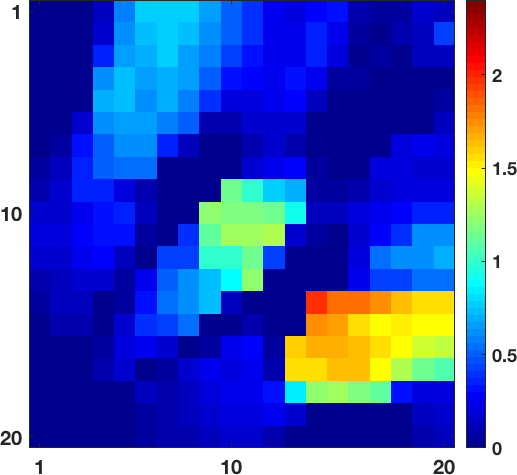}
\label{fig:9b}}
\hfil
\subfloat[Scale 3: RMSE=0.4220]{\includegraphics[width=2in]{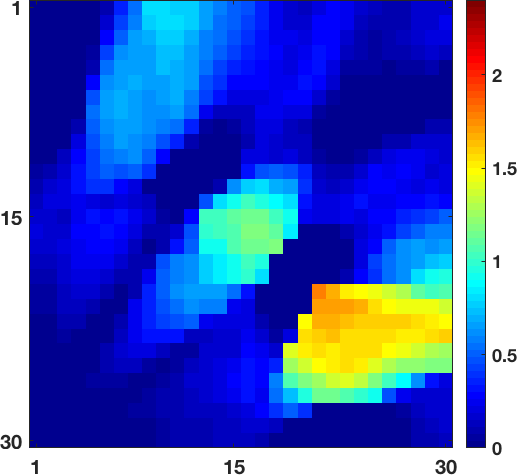}
\label{fig:9c}}
\vfil
\subfloat[Scale 4: RMSE=0.3209]{\includegraphics[width=2in]{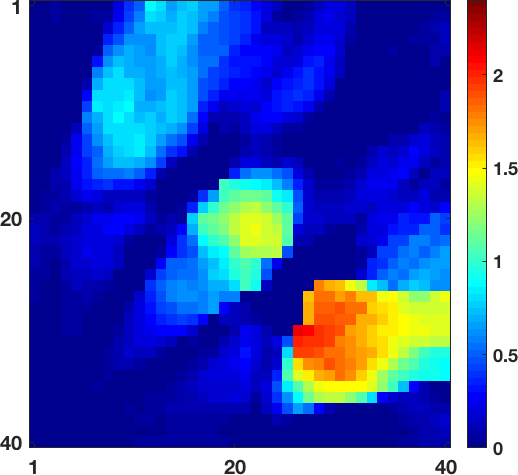}
\label{fig:9d}}
\hfil
\subfloat[Scale 5: RMSE=0.3137]{\includegraphics[width=2in]{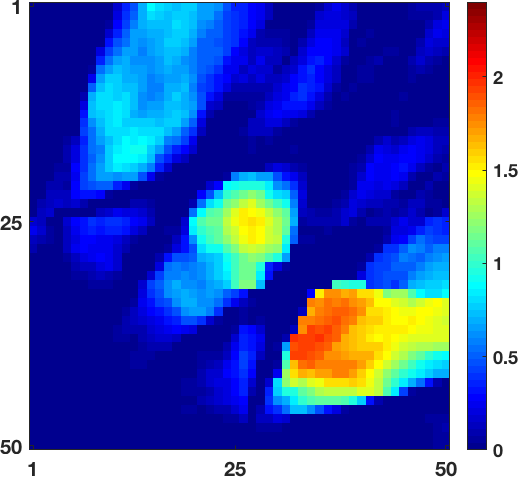}
\label{fig:9e}}

 \caption{ Density reconstruction results using attenuation data alone for the Phantom-II obtained at each scale of processing for the first iteration of the algorithm. Subplots (a)-(e) show the density reconstruction results for $5$ different grid sizes from $10\times 10$ to $50\times 50$.}
 \label{fig:9}
\end{figure*}

 Density reconstruction results and associated RMSE for attenuation-only data are shown in Fig. \ref{fig:8} and Fig. \ref{fig:9} for the first and second phantoms respectively. These images indicate that attenuation-only data, while providing reconstructions whose amplitudes are in the right range, suffer from significant artifacts making clear identification of the distinct regions in the scene virtually impossible.  
 
\par
\begin{figure*}[!t]
\centering
\subfloat[Scale 1: RMSE=0.7524]{\includegraphics[width=2in]{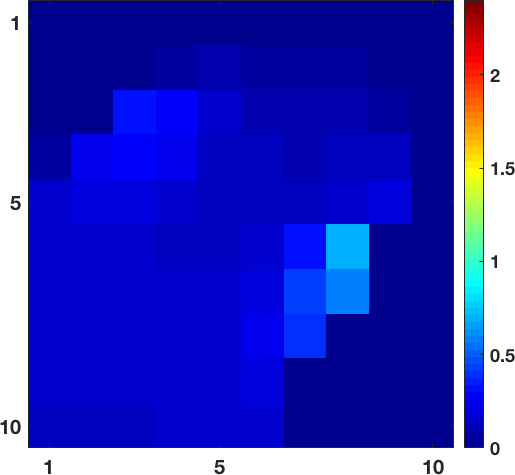}
\label{fig:10a}}
\hfil
\subfloat[Scale 2: RMSE=0.4020]{\includegraphics[width=2in]{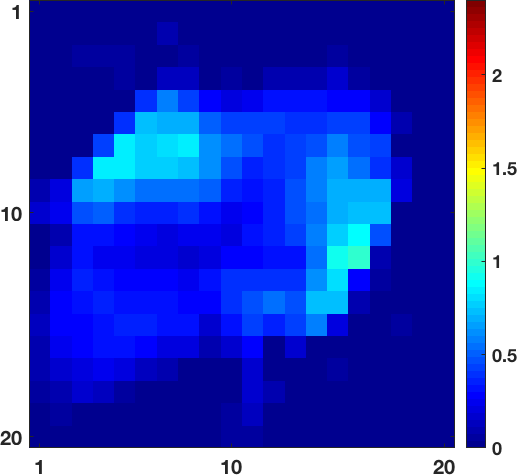}
\label{fig:10b}}
\hfil
\subfloat[Scale 3: RMSE=0.3765]{\includegraphics[width=2in]{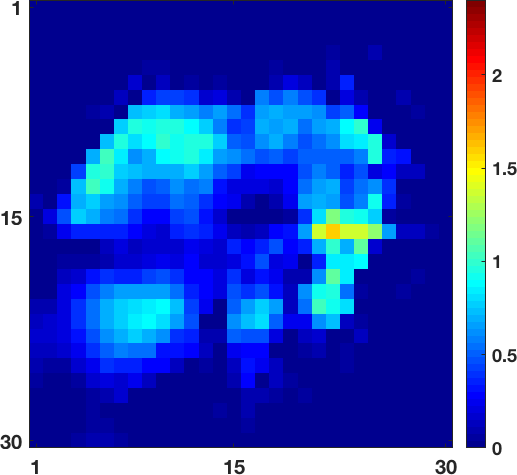}
\label{fig:10c}}
\vfil
\subfloat[Scale 4: RMSE=0.3194]{\includegraphics[width=2in]{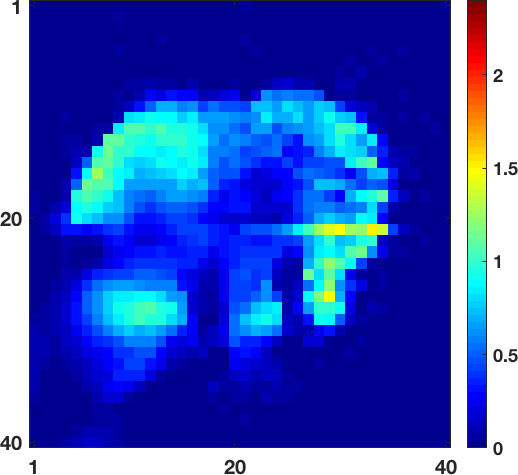}
\label{fig:10d}}
\hfil
\subfloat[Scale 5: RMSE=0.2989]{\includegraphics[width=2in]{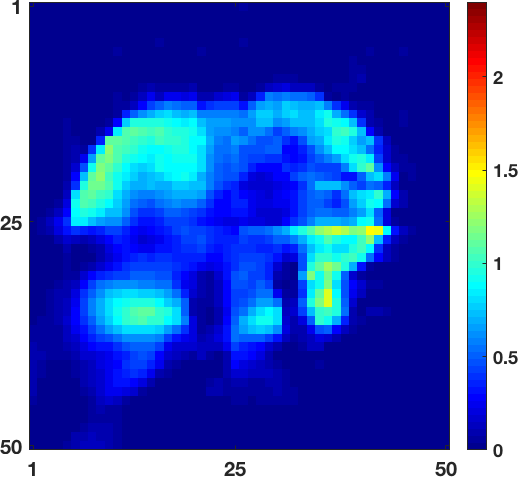}
\label{fig:10e}}
 \caption{Density reconstruction results with only scatter data for Phantom-I for each scale of processing at the first iteration of the algorithm. Density reconstruction using only scatter data is successful in recovering the structure of the object compared to attenuation only data but has large relative mean-squared error due to inaccuracy in the overall amplitude. Subplots (a)-(e) show the density reconstruction results for $5$ different grid sizes from $10\times 10$ to $50\times 50$.}
 \label{fig:10}
\end{figure*}

\begin{figure*}[!t]
\centering
\subfloat[Scale 1: RMSE=0.6283]{\includegraphics[width=2in]{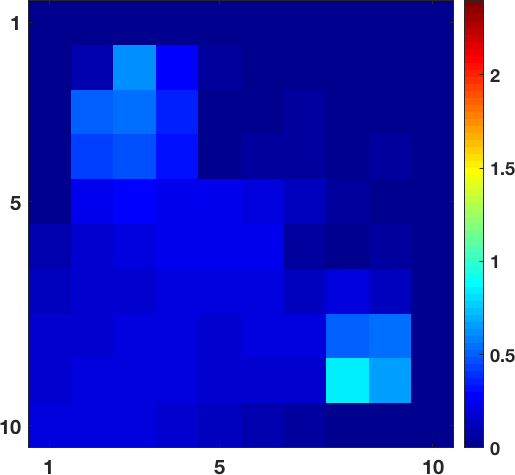}
\label{fig:11a}}
\hfil
\subfloat[Scale 2: RMSE=0.5776]{\includegraphics[width=2in]{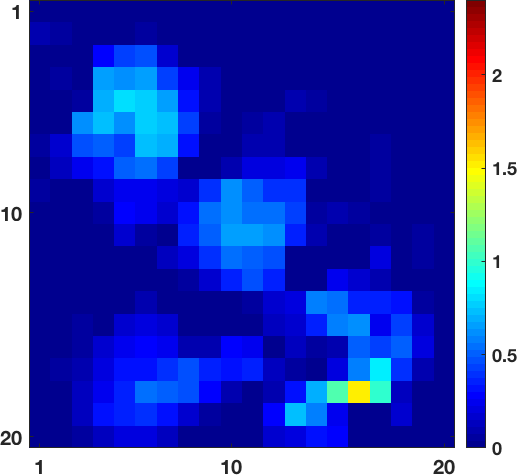}
\label{fig:11b}}
\hfil
\subfloat[Scale 3: RMSE=0.5006]{\includegraphics[width=2in]{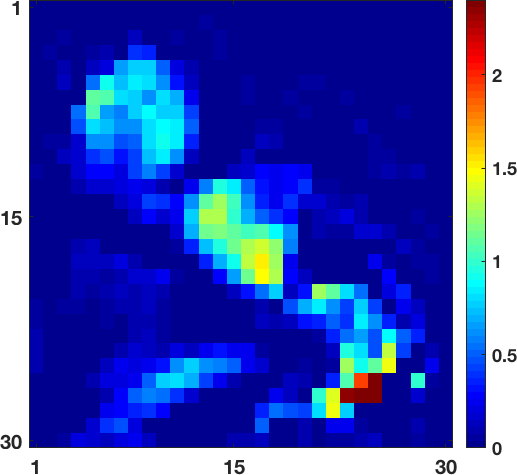}
\label{fig:11c}}
\vfil
\subfloat[Scale 4: RMSE=0.4657]{\includegraphics[width=2in]{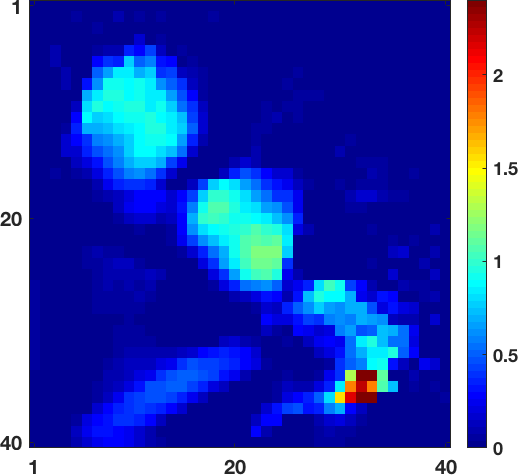}
\label{fig:11d}}
\hfil
\subfloat[Scale 5: RMSE=0.4511]{\includegraphics[width=2in]{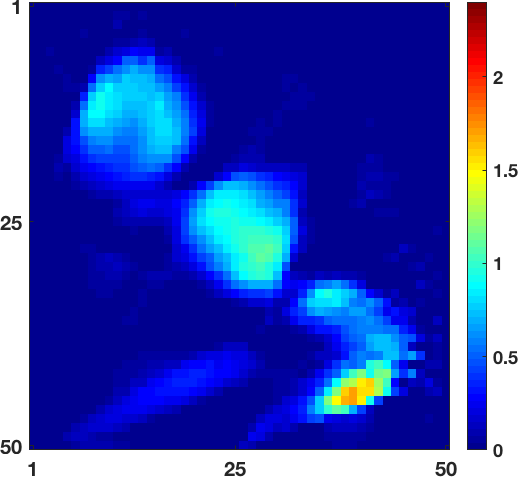}
\label{fig:11e}}

 \caption{Density reconstruction results with only scatter data for Phantom-II for each scale of processing at the first iteration of the algorithm. Subplots (a)-(e) show the density reconstruction results for $5$ different grid sizes from $10\times 10$ to $50\times 50$.}
 \label{fig:11}
\end{figure*}

 Density reconstruction images and RMSE with scattering-only data are demonstrated in Fig. \ref{fig:10} and Fig. \ref{fig:11}. Scattering-only data provides reconstructions where the structure of the objects are better recovered and the artifacts are reduced significantly, however the amplitudes are not completely in the right range relative to attenuation-only data.  
By comparing Fig. \ref{fig:8} (a)-(e) and Fig. \ref{fig:10} (a)-(e) for the first phantom, the shape of the elephant is better recovered in scattering-only data and RMSE at each scale is smaller relative to attenuation-only data case. From Fig. \ref{fig:9}(a)-(e), for the second phantom the attenuation-only density reconstructions contain artifacts and noise around and along the objects so the structure of the objects are not well recovered. On the other hand, the reconstruction obtained using only scatter data contains fewer artifacts and the shape of the objects is generally better defined. We do note that the RMSE for the attenuation-only data is still smaller than that obtained using the scatter data as the absolute amplitudes of the objects are more accurate for the attenuation data even if their precise geometry is worse. 
  
\par
\begin{figure*}[!t]
\centering
\subfloat[Scale 1: RMSE=0.1989]{\includegraphics[width=2in]{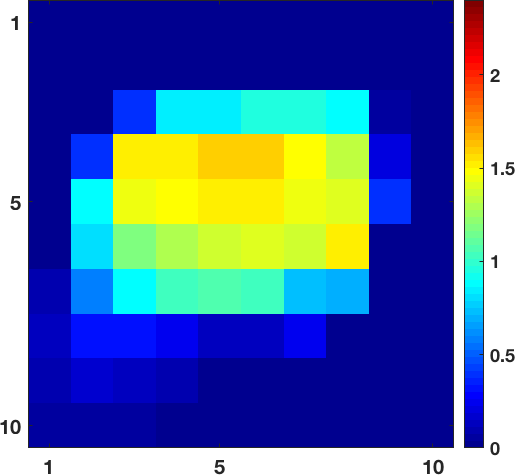}
\label{fig:12a}}
\hfil
\subfloat[Scale 2: RMSE=0.1724]{\includegraphics[width=2in]{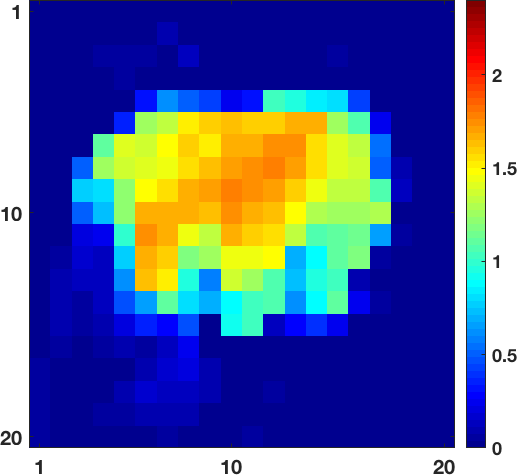}
\label{fig:12b}}
\hfil
\subfloat[Scale 3: RMSE=0.1476]{\includegraphics[width=2in]{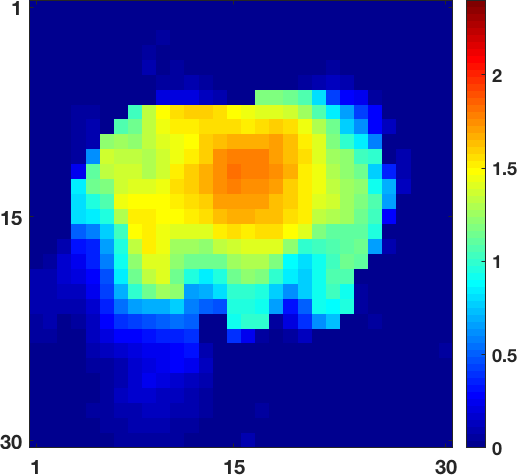}
\label{fig:12c}}
\vfil
\subfloat[Scale 4: RMSE=0.1370]{\includegraphics[width=2in]{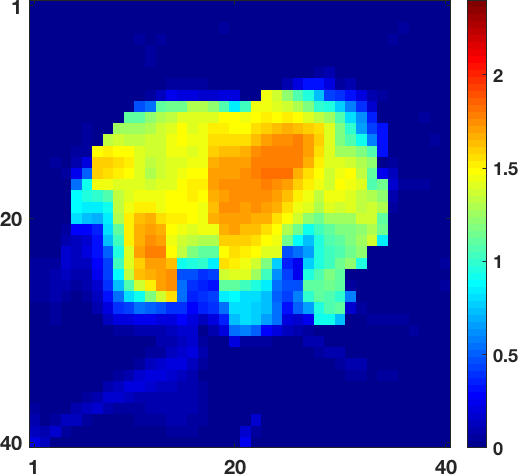}
\label{fig:12d}}
\hfil
\subfloat[Scale 5: RMSE=0.0811]{\includegraphics[width=2in]{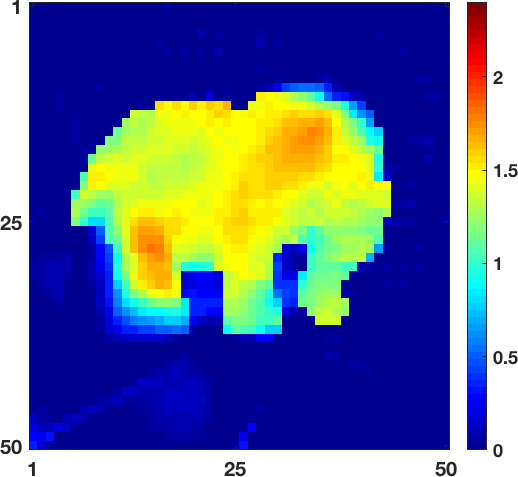}
\label{fig:12e}}
 \caption{Density reconstruction results with both attenuation and scatter data for Phantom-I for each scale of processing at the first iteration of the algorithm. The combination of datasets improves the performance of the density reconstruction by taking advantage of scatter data in recovering the structure of the object and attenuation data in increasing the accuracy of the reconstructed amplitudes. Subplots (a)-(e) show density reconstruction results for $5$ different grid sizes from $10\times 10$ to $50\times 50$.}
 \label{fig:12}
\end{figure*}

\begin{figure*}[!t]
\centering
\subfloat[Scale 1: RMSE=0.5307]{\includegraphics[width=2in]{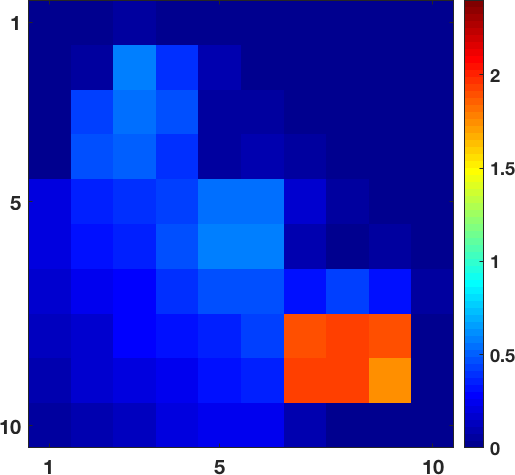}
\label{fig:13a}}
\hfil
\subfloat[Scale 2: RMSE=0.2183]{\includegraphics[width=2in]{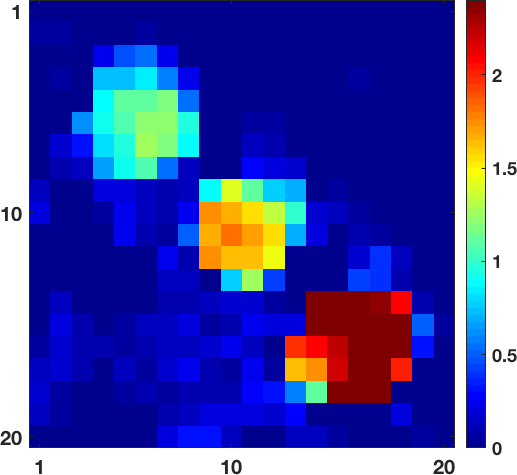}
\label{fig:13b}}
\hfil
\subfloat[Scale 3: RMSE=0.1281]{\includegraphics[width=2in]{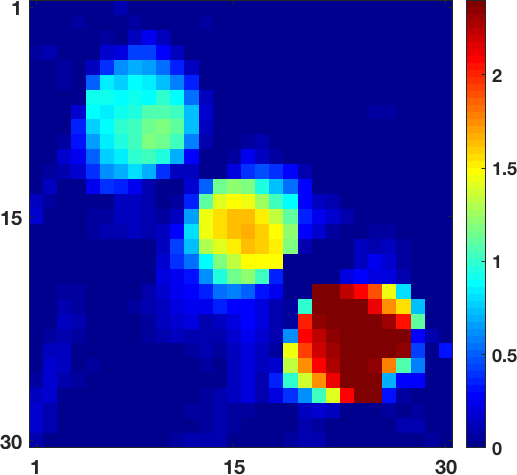}
\label{fig:13c}}
\hfil
\subfloat[Scale 4: RMSE=0.1134]{\includegraphics[width=2in]{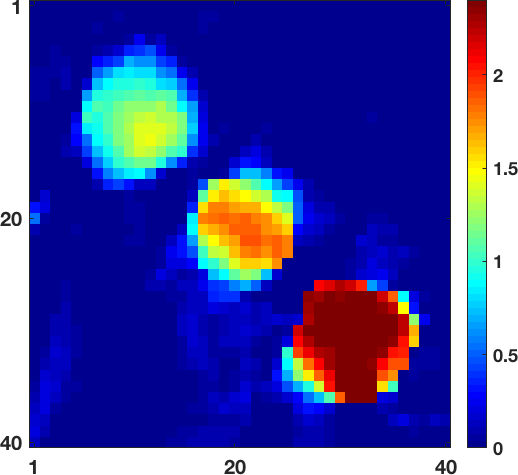}
\label{fig:13d}}
\hfil
\subfloat[Scale 5: RMSE=0.0941]{\includegraphics[width=2in]{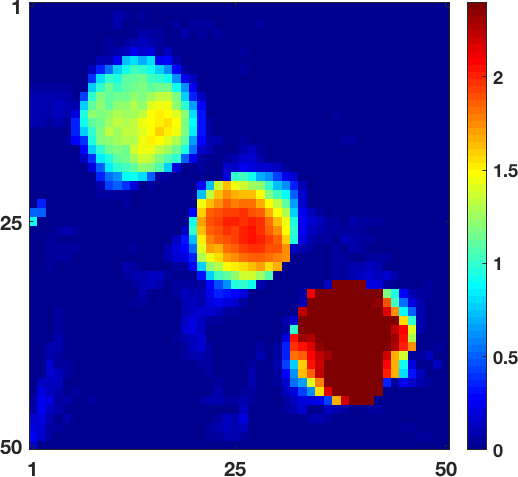}
\label{fig:13e}}

 \caption{Density reconstruction results with both attenuation and scatter data for Phantom-II for each scale of processing at the first iteration of the algorithm. Subplots (a)-(e) show the density reconstruction results for $5$ different grid sizes from $10\times 10$ to $50\times 50$.}
 \label{fig:13}
\end{figure*}

 Density reconstructions derived from combination of both attenuation and scattering information at the first iteration are shown in Fig. \ref{fig:12} and Fig. \ref{fig:13}. These images clearly demonstrate the advantages (both quantitative and qualitative) of employing both types of data. Specifically, both the geometric structure of the objects as well as the pixel-by-pixel estimates of the density value are improved in the latter optimization compared to the previous examples. For example, concavities between the elephant's legs and the front leg and trunk are better resolved and we are able to distinguish both the geometries of the three separate shapes as well as the material properties in the second phantom. Finally, we see far fewer background artifacts and note that the RMSE at the end of the multi-scale process is reduced by $74.43\%$ and $70\%$ relative to the attenuation-only reconstructions for the first and second phantom respectively and $72.87\%$ and $79.14\%$ relative to the scatter-only reconstructions.
\par Fig. \ref{fig:8}-Fig. \ref{fig:13} also provide evidence of the utility of the multi-scale approach. The multi-scale approach starting from the grid with the size of $10\times 10$ ending with the grid of the size of $50\times 50$ is applied to both of the phantoms. The results of different scales for the three different data settings are shown in Fig. \ref{fig:8}, Fig. \ref{fig:10} and Fig. \ref{fig:12} for the first phantom and  Fig. \ref{fig:9}, Fig. \ref{fig:11} and Fig. \ref{fig:13} for the second phantom respectively. The approach performed well using a spatially constant initial guess for the density and zero for the photoelectric absorption.  Specifically, where both absorption and scatter data are employed we see a monotonic decrease in the RMSE as well as qualitative improvements as we refine the scale. \par

\begin{figure*}[!t]
\centering
\subfloat[RMSE=0.3478]{\includegraphics[width=2in]{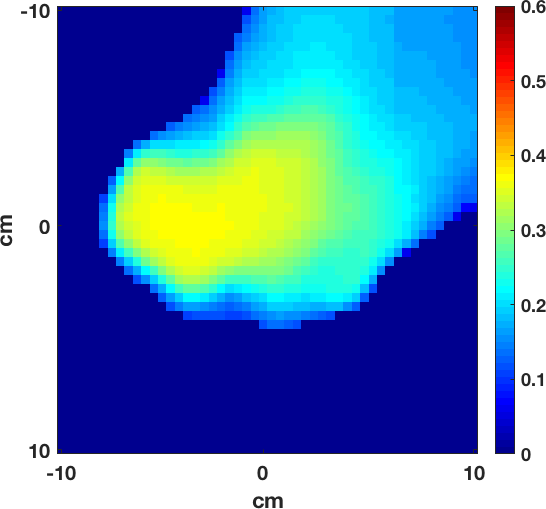}
\label{fig:14a}}
\hfil
\subfloat[RMSE=0.1421]{\includegraphics[width=2in]{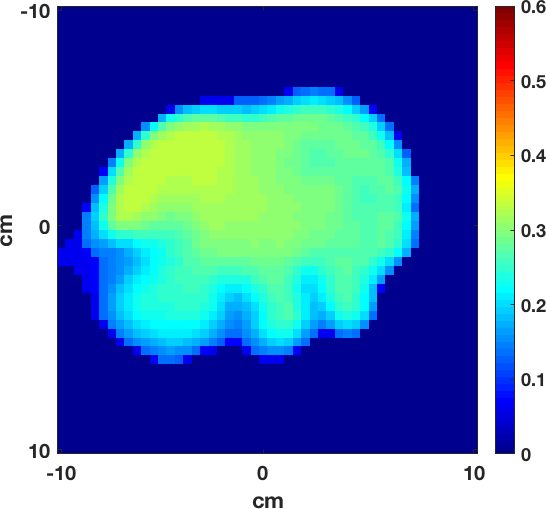}
\label{fig:14b}}
\hfil
\subfloat[RMSE=0.1020]{\includegraphics[width=2in]{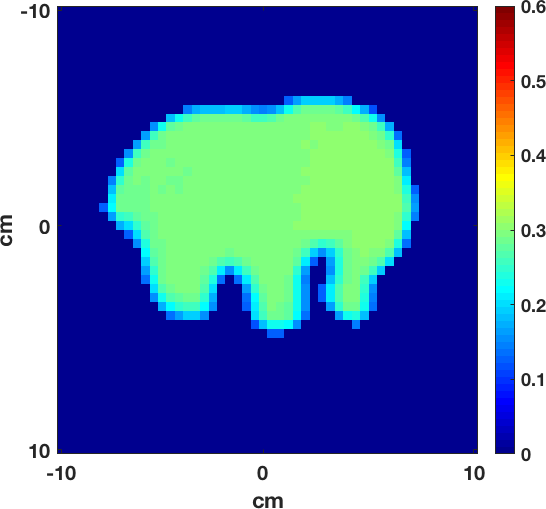}
\label{fig:14c}}
\vfil
\subfloat[RMSE=0.1201]{\includegraphics[width=2in]{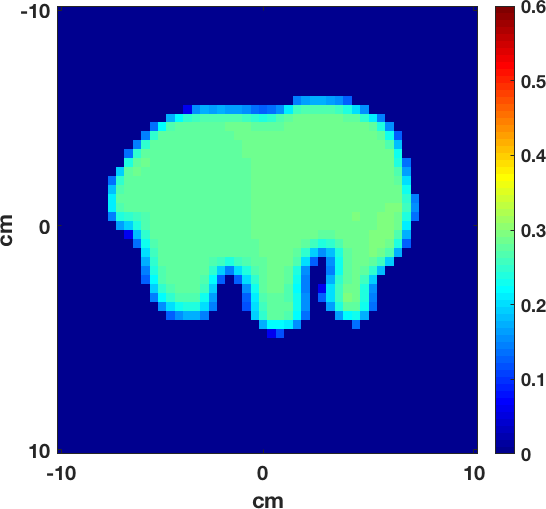}
\label{fig:14d}}
\hfil
\subfloat[RMSE=0.0675]{\includegraphics[width=2in]{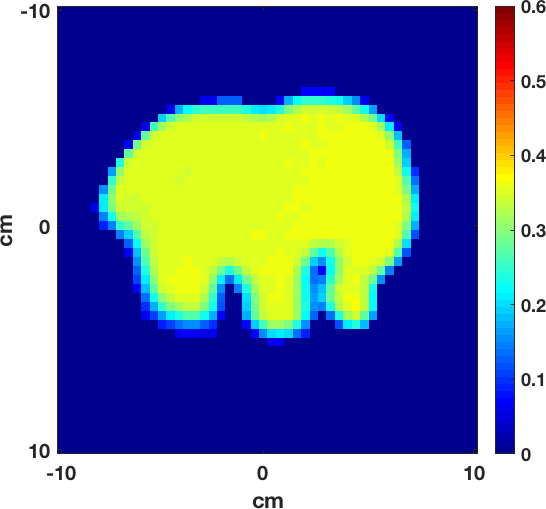}
\label{fig:14e}}
 \caption{Recovery of photoelectric map at first iteration of the algorithm for Phantom-I.  In (a) only attenuation data is used for estimating $\hat{\boldsymbol\rho}_1$ and the photoelectric is also estimated using only attenuation data.  In (b) both density and photoelectric are estimated using only scatter data. In (c) photoelectric is estimated with attenuation-only data while we use as $\hat{\boldsymbol\rho}_1$ the reconstruction in Fig. \ref{fig:12}(e) obtained using both scatter and attenuation data. In (d) photoelectric is estimated with only scatter data while employing the density estimated from both datasets. In (e) both density and photoelectric are estimated using both datasets. The quantitative measure RMSE confirms that the combination of scattering and attenuation datasets in density reconstruction increases the accuracy of photoelectric reconstruction.}
 \label{fig:14}
\end{figure*}

\begin{figure*}[!t]
\centering
\subfloat[RMSE=0.7074]{\includegraphics[width=2in]{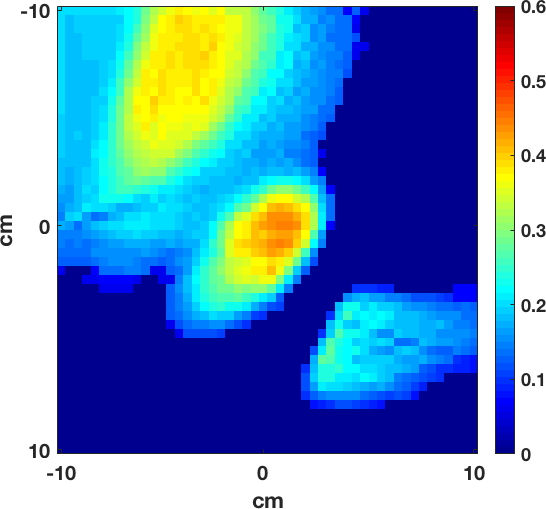}
\label{fig:15a}}
\hfil
\subfloat[RMSE=0.6448]{\includegraphics[width=2in]{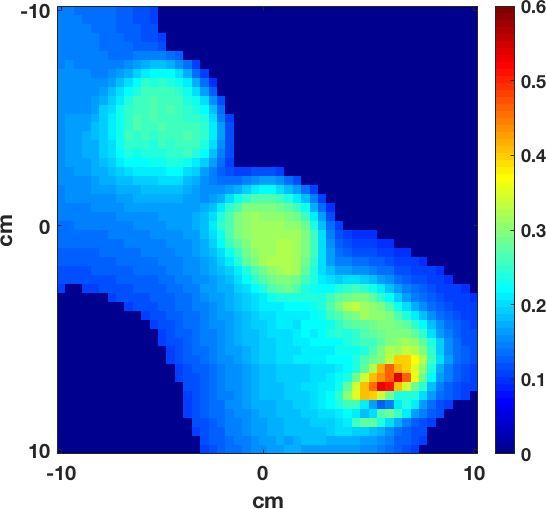}
\label{fig:15b}}
\hfil
\subfloat[RMSE=0.1677]{\includegraphics[width=2in]{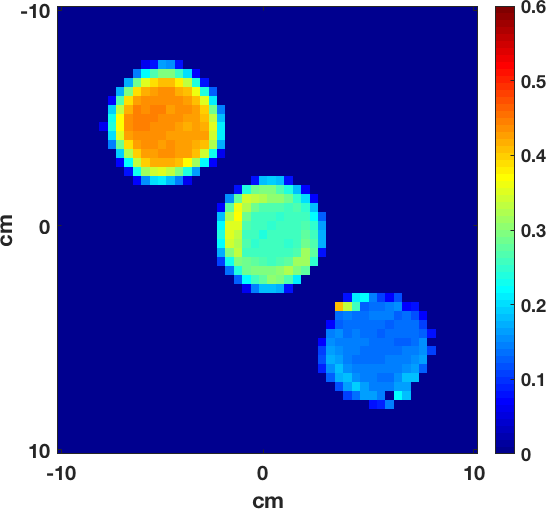}
\label{fig:15c}}
\vfil
\subfloat[RMSE=0.1504]{\includegraphics[width=2in]{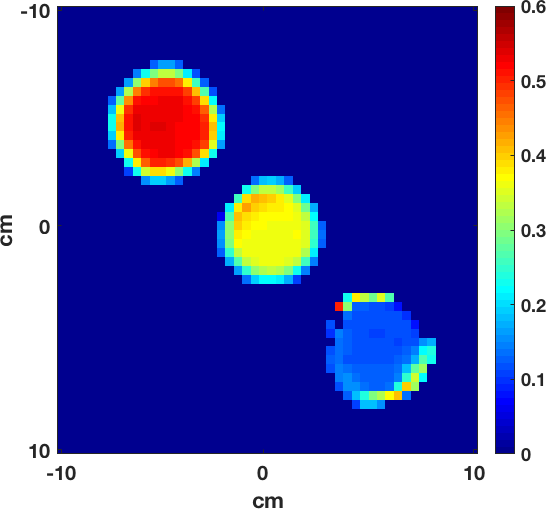}
\label{fig:15d}}
\hfil
\subfloat[RMSE=0.1479]{\includegraphics[width=2in]{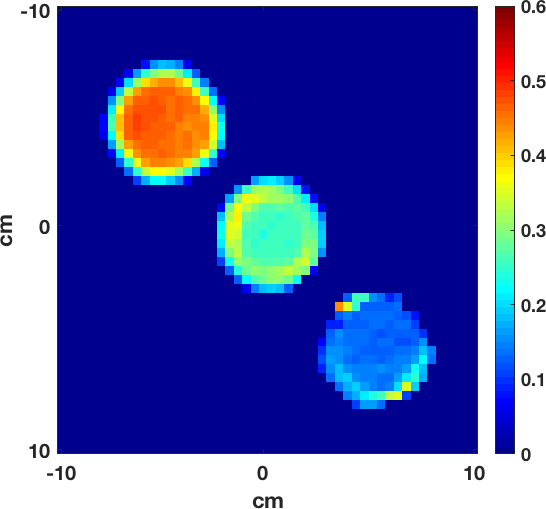}
\label{fig:15e}}
 \caption{Recovery of photoelectric map at first iteration of the algorithm for Phantom-II. In (a) only attenuation data is used for estimating $\hat{\boldsymbol\rho}_1$ and the photoelectric is also estimated using only attenuation data. In (b) both density and photoelectric are estimated using only scatter data. In (c) photoelectric is estimated with attenuation-only data while we use as $\hat{\boldsymbol\rho}_1$ the reconstruction in Fig. \ref{fig:13}(e) obtained using both scatter and attenuation data. In (d) photoelectric is estimated with only scatter data while employing the density estimated from both datasets. In (e) both density and photoelectric are estimated using both datasets. The quantitative measure RMSE confirms that the combination of scattering and attenuation datasets in density reconstruction increases the accuracy of photoelectric reconstruction.}
 \label{fig:15}
\end{figure*}

Having examined the utility of different data types on our ability to recover mass density, we now turn our attention to mapping the photoelectric attenuation coefficient. As in the case of density, we wish to explore the impact of attenuation-only, scattering-only and combination of both datasets in reconstructing photoelectric coefficient at first iteration. 
Since photoelectric reconstruction is very sensitive to noise our ability to recover this quantity is very dependent to the quality of density reconstruction \cite{r15}, \cite{r14}. To investigate the effect of density estimation on photoelectric reconstruction when only attenuation data are used for $\hat{\boldsymbol\rho}_1$ in (\ref{eq:32}) we use the attenuation-only density reconstruction in Fig. \ref{fig:8}(e) for the first phantom and Fig. \ref{fig:9}(e) for the second. The resulting estimates of photoelectric coefficient for this case are shown in Fig. \ref{fig:14}(a) and Fig. \ref{fig:15}(a) respectively. The same procedure as applied to scattering-only data yields the results in Fig. \ref{fig:14}(b)  and Fig. \ref{fig:15}(b). 
In both cases, the errors associated with the density initialization lead to relatively poor recovery of photoelectric. 
Next, for both attenuation-only and scattering-only photoelectric optimization process we use for $\hat{\boldsymbol\rho}_1$ the more accurate density estimate provided in Fig. \ref{fig:12}(e) and Fig. \ref{fig:13}(e) where both scattering and attenuation data are used in the initial recovery of mass density.  
The results are shown in Fig. \ref{fig:14}(c)  and Fig. \ref{fig:15}(c) and Fig. \ref{fig:14}(d) and Fig. \ref{fig:15}(d) for each dataset respectively. In this case, photoelectric reconstructions are much more accurate and RMSE has decreased significantly. These results provide strong evidence that the accuracy in density estimation plays a critical role in photoelectric reconstruction.
\par Finally, for the case where we have used both attenuation and scatter data for photoelectric reconstruction, we have initialized  $\hat{\boldsymbol\rho}_1$  with density estimation obtained using the combination of attenuation and scattering density reconstruction with the results shown in Fig. \ref{fig:14}(e)  and Fig. \ref{fig:15}(e). Comparing the latter case with the previous cases shows that combination of both datasets improves the accuracy of photoelectric reconstruction significantly, since the density reconstruction derived from both dataset. 
 
\begin{figure*}[!ht]
\centering
\subfloat[RMSE=0.0166]{\includegraphics[width=2in]{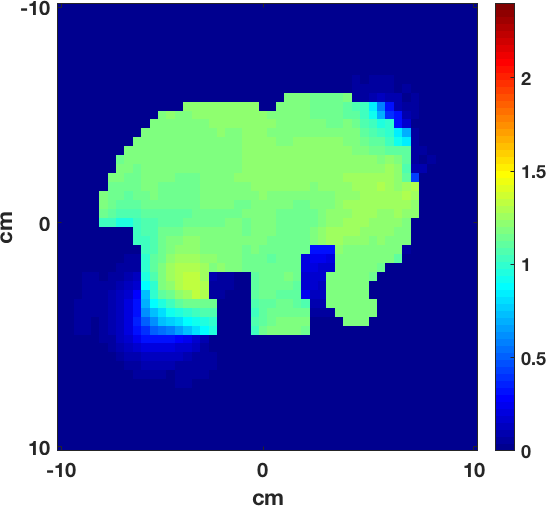}
\label{fig:16a}}
\hfil
\subfloat[RMSE=0.0383]{\includegraphics[width=2in]{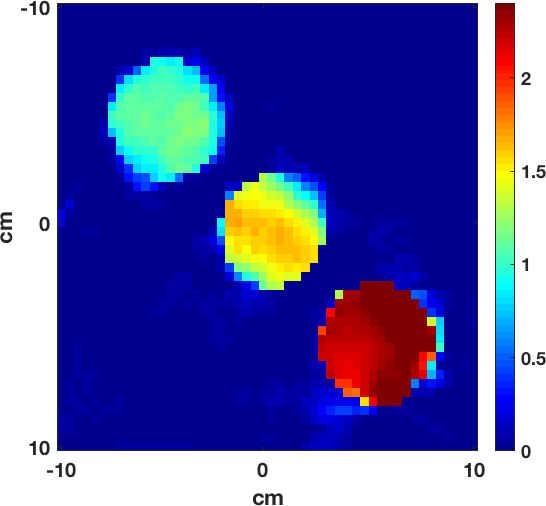}
\label{fig:16b}}
\vfil
\subfloat[RMSE=0.0091]{\includegraphics[width=2in]{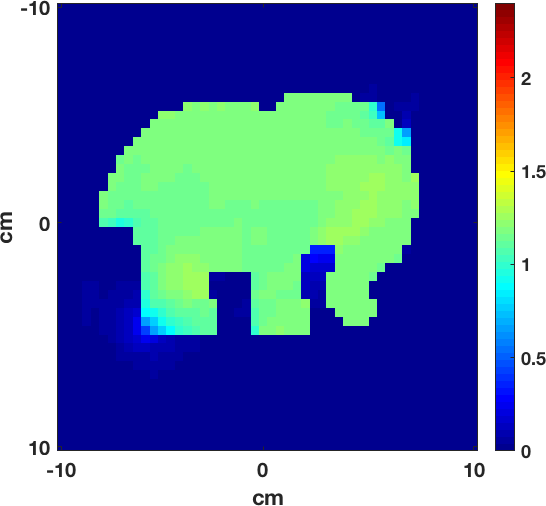}
\label{fig:16c}}
\hfil
\subfloat[RMSE=0.0251]{\includegraphics[width=2in]{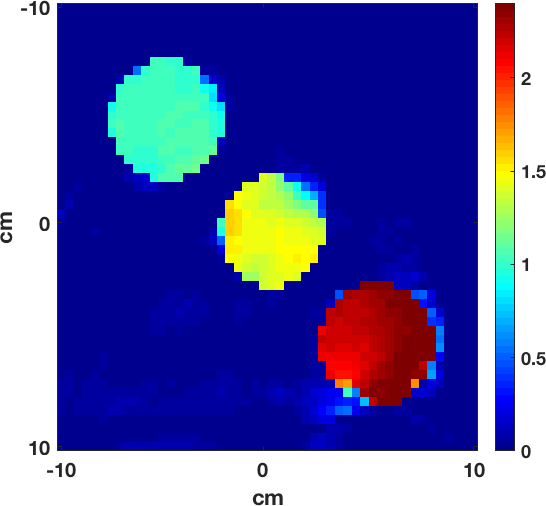}
\label{fig:16d}}

\caption{Density reconstruction for both of the phantoms with associated RMSE at second iteration are shown in (a) and (b) while (c) and (d) show the third iteration results.}
 \label{fig:16}
\end{figure*}

\begin{figure*}[!ht]
\centering
\subfloat[RMSE=0.0645]{\includegraphics[width=2in]{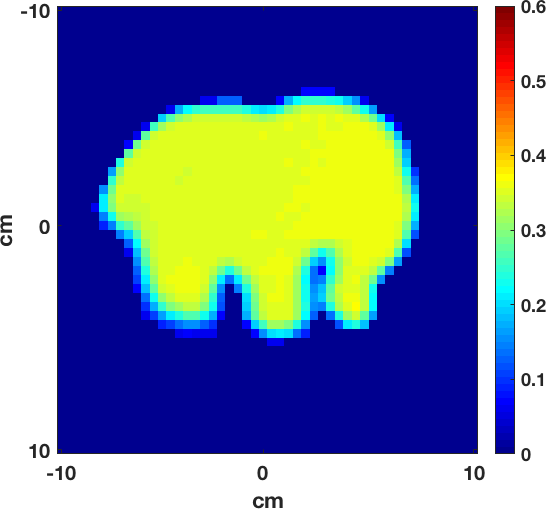}
\label{fig:17a}}
\hfil
\subfloat[RMSE=0.0953]{\includegraphics[width=2in]{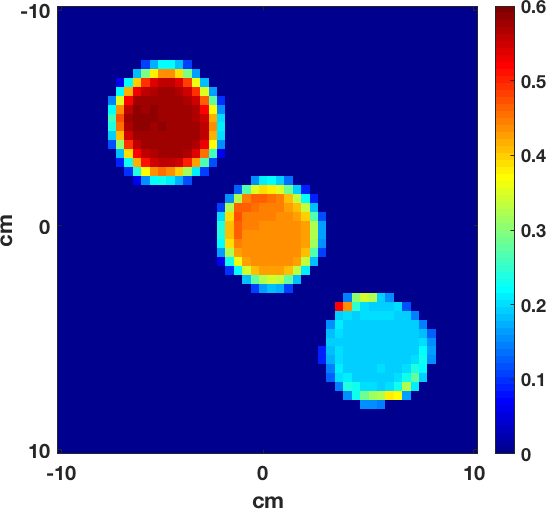}
\label{fig:17b}}
\vfil
\subfloat[RMSE=0.0589]{\includegraphics[width=2in]{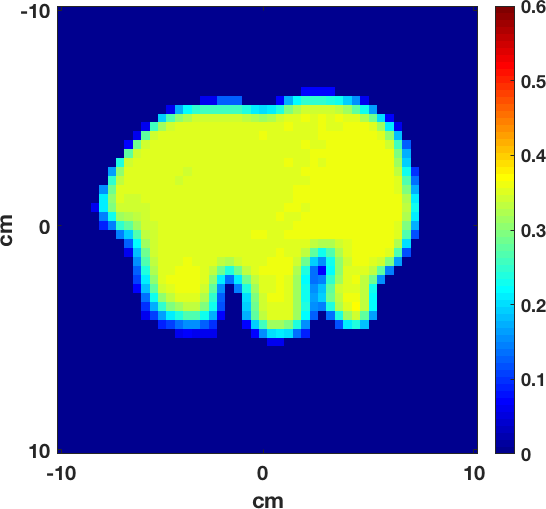}
\label{fig:17c}}
\hfil
\subfloat[RMSE=0.0818]{\includegraphics[width=2in]{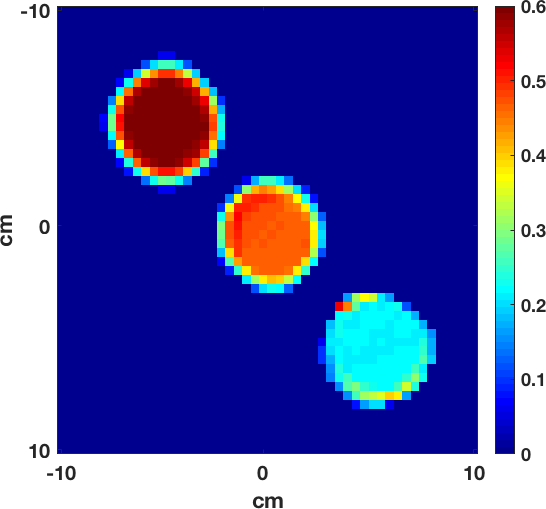}
\label{fig:17d}}

\caption{Photoelectric coefficient reconstruction images and RMSE measure  for the second iteration of the algorithm are shown by (a) and (b) while (c) and (d) show the third iteration reconstructions.}
 \label{fig:17}
\end{figure*}

In Figures Fig. \ref{fig:16} and  Fig. \ref{fig:17} we display the density and photoelectric reconstructions using both data sets for the second and third iterations of  the algorithm.  While the results for even the first iteration were rather good especially given the limited view nature of the problem, we do see both quantitative and qualitative improvements from the continued processing. Indeed, with $\epsilon=0.01$, the first convergence criterion in  (\ref{eq:42}) is achieved for $n=3$.

\par
\begin{figure*}[!ht]
\centering
\subfloat[]{\includegraphics[width=2in]{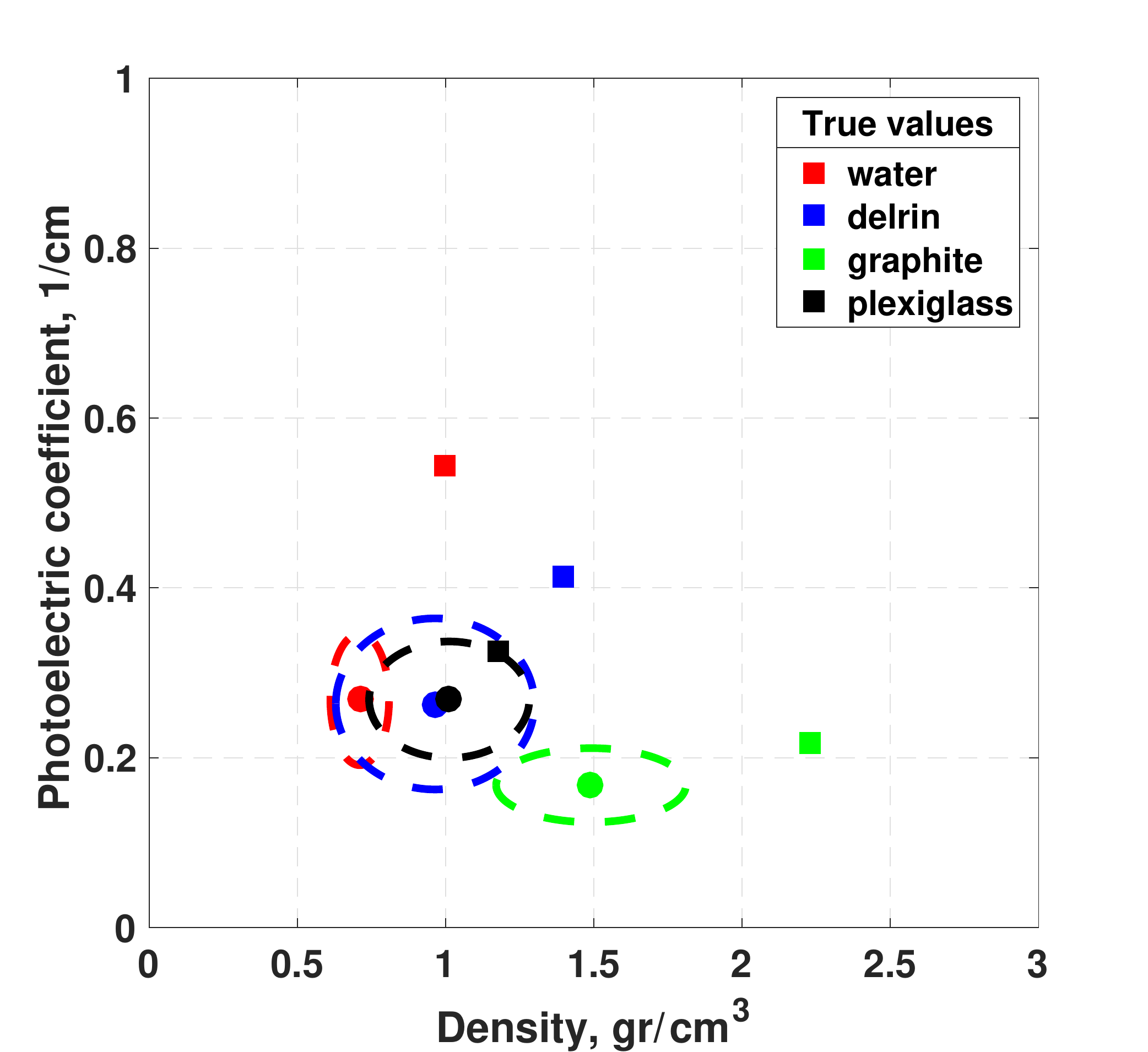}
\label{fig:18a}}
\hfil
\subfloat[]{\includegraphics[width=2in]{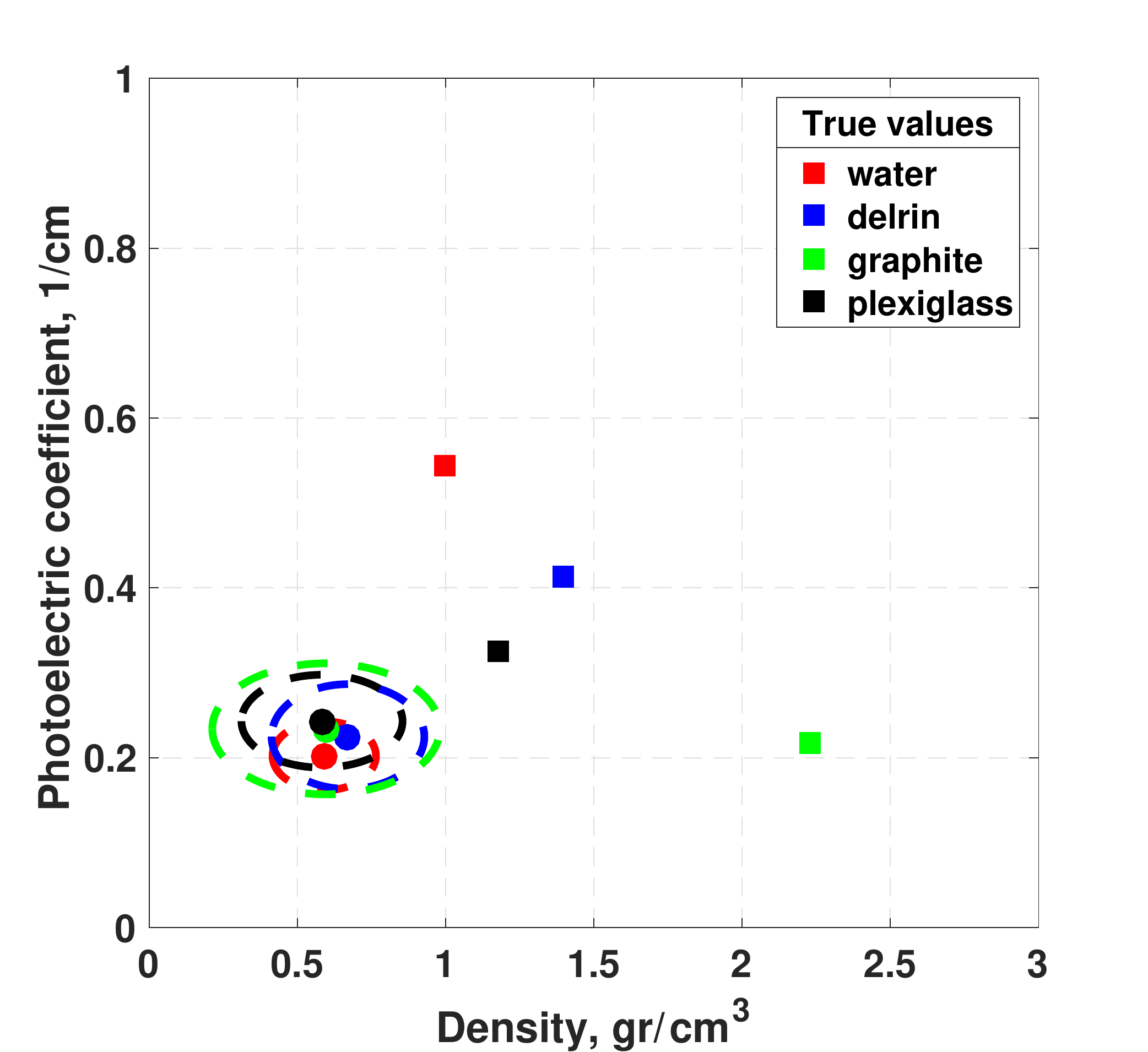}
\label{fig:18b}}
\hfil
\subfloat[]{\includegraphics[width=2in]{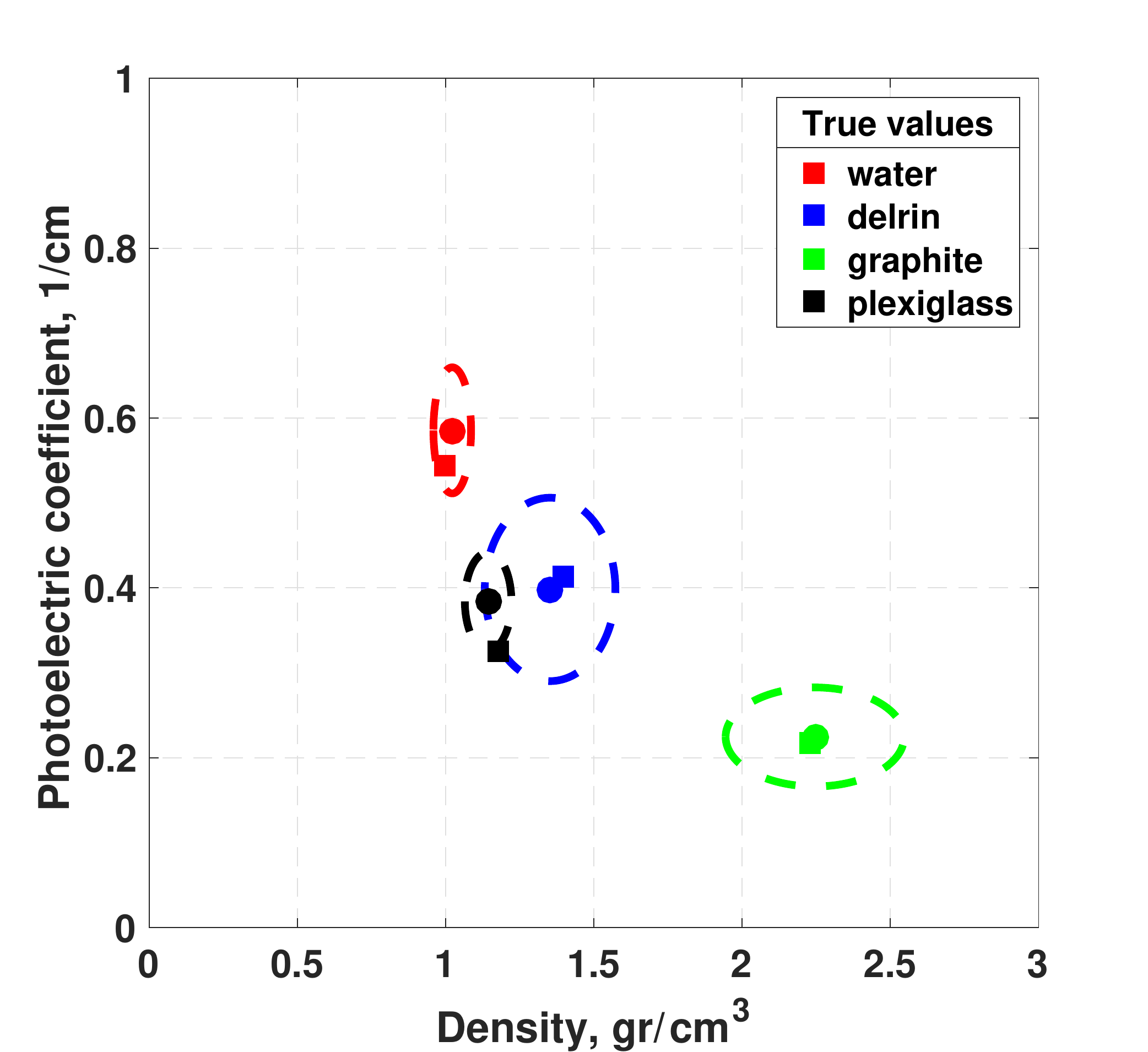}
\label{fig:18c}}
\caption{Material characterization uncertainty ellipses using attenuation-only, scattering-only and scattering-attenuation datasets. The true value of the four different objects are shown by `$\blacksquare$' in different colors. The reconstructed objects for each datasets are segmented and the mean and standard deviation are computed to generate the uncertainty ellipses. The mean of each segment is shown by `$\protect\newsymbol$'. (a) Attenuation-only dataset uncertainty results. (b) Scatter-only dataset uncertainty results.  (c) Scatter and attenuation dataset uncertainty results.}
 \label{fig:18}
\end{figure*}

 Finally, we examine the performance of the proposed method in terms of quantitative material characterization. The objects  are manually segmented in density and photoelectric images, and  the mean and standard deviation of the pixels belonging to the individual segment are calculated. Uncertainty ellipses are plotted for each object as an ellipse centered by the mean in density and photoelectric images and one standard deviation for semi-major/minor axis. The ellipses for four different material are plotted in Fig. \ref{fig:18} comparing attenuation-only, scattering-only and combination of both datasets. In either of Fig. \ref{fig:18}(a)  and Fig. \ref{fig:18}(b) these clouds are not centered around the true value of the associated material and have higher standard deviation while in Fig. \ref{fig:18}(c) which shows the results of combination of both datasets, the mean of each segment is close to the true value of that segment and the standard deviation at each direction is reduced. 

  \section{Conclusion}
  \label{sec:V}
In this paper we have demonstrated empirically the advantages obtained by fusing energy-resolved attenuation and Compton scatter data for the joint recovery of mass density and photoelectric absorption properties and subsequent quantitative materials characterization in the context of severely limited view geometries. After developing both the underlying physical model and associated numerical implementation for the Compton scatter process, we propose a variational scheme for estimation of the two material properties of interest. We have proposed a cyclic descent method for reconstruction of density and photoelectric images where at the first iteration we have applied a multi-scale approach to estimate density without requiring any prior knowledge about the objects. We have also shown that with properly choosing the regularization method the quality of reconstruction will be increased. In density reconstruction we have applied an iterative edge-preserving method which is successful in capturing the details of the objects. We have shown that the quality of density reconstruction has a direct impact in photoelectric stabilized reconstruction which is accomplished with NLM regularization and reconstruction of density with combination of scattering and attenuation data. In terms of material characterization we have also analyzed the performance of the system by plotting uncertainty ellipses. Combining both sets of data allows us to characterize different materials with higher certainty than can be obtained using either data set alone. 
\par In future work, we will modify the noise statistics to be more compatible with the nature of the photon-counting model. In the current work we are ignoring out-of-plane scattering while there is a fraction of photons that are captured out of plane. By adding out-of-plane detectors, we could capture $3D$ scattering that could improve the performance of the system. However, this also leads to a coupled $3D$ inversion problem which would impose severe computational loads. Another direction for future work is efficiently using fast and parallel algorithms in order to achieve a real-time reconstruction algorithm, which plays an important role in real-time applications like the baggage screening.
\par Another topic for future work is development of an improved and efficient method for choosing the regularization parameters to guarantee the best construction results for both density and photoelectric. 
\appendices
\section{Calculating Jacobian Matrix}
To reconstruct photoelectric image the first derivative of objective function introduced in (\ref{eq:32}) is required. It can be facilitated by rewriting (\ref{eq:32}) as
\begin{equation}
	F(\textbf{p})=w_1\| \textbf{g}_{C}-\textbf{K}_{C}(\boldsymbol\rho_n,\textbf{p})\boldsymbol\rho_n\|_{2}^{2}+w_2\| \textbf{g}_{A}-\textbf{K}_{A,\rho}\boldsymbol\rho_n-\textbf{K}_{A,p}\textbf{p}\|_{2}^{2}+R_{p}(\textbf{p}|\textbf{I}^{ref})=\textbf{f}(\textbf{p})^T\textbf{f}(\textbf{p})
\label{eq:43}
\end{equation}
where $\textbf{f}(\textbf{p})=[\textbf{f}_a(\textbf{p});\textbf{f}_c(\textbf{p});\textbf{f}_r(\textbf{p})]$  includes the data mismatch for attenuation and scattering and regularization terms respectively given as 
\begin{equation}
	\textbf{f}_a(\textbf{p})=  \sqrt{w_2}\left[\textbf{g}_{A}-\textbf{K}_{A,\rho}\boldsymbol\rho_n-\textbf{K}_{A,p}\textbf{p}\right] 
\label{eq:44}
\end{equation}
\begin{equation}
	\textbf{f}_c(\textbf{p})=  \sqrt{w_1}\left[\textbf{g}_{C}-\textbf{K}_{C}(\boldsymbol\rho_n,\textbf{p})\boldsymbol\rho_n\right] 
\label{eq:45}
\end{equation}
\begin{equation}
	\textbf{f}_r(\textbf{p})=\sqrt{\lambda_p}(\textbf{I}-\textbf{W})\textbf{p}
\label{eq:46}
\end{equation}

The Jacobian matrix with respect to $\textbf{p}$ can be derived analytically by calculating first derivative of $\textbf{f}_a(\textbf{p})$,$\textbf{f}_c(\textbf{p})$ and $\textbf{f}_r(\textbf{p})$ as  
\begin{equation}
	\textbf{J}=\left[\frac{\partial \textbf{f}(\textbf{p})}{\partial{\textbf{p}}}\right]=
	\left[\frac{\partial{\textbf{f}_a(\textbf{p})}}{\partial{\textbf{p}}};\frac{\partial{\textbf{f}_c(\textbf{p})}}{\partial{\textbf{p}}};\frac{\partial{\textbf{f}_r(\textbf{p})}}{\partial{\textbf{p}}}\right]
\label{eq:47}
\end{equation}
with 
\begin{equation}
	\frac{\partial{\textbf{f}_r(\textbf{p})}}{\partial{\textbf{p}}}=\sqrt{\lambda_p}(\textbf{I}-\textbf{W})
\label{eq:48}
\end{equation}
for NLM regularization scheme. The Jacobian matrix of the attenuation mismatch term can be found in \cite{r14}. 
For the scattering data mismatch term, the  $j$-th row of the Jacobian matrix associated with the forward model is
\begin{equation}
\begin{split}
	\left[\frac{\partial{f_c(p)}}{\partial{p}}\right]_{j} &=\sqrt{w_2}\left[\frac{\partial{K_{C}(\rho=\rho_n,p)\rho_n}}{\partial{p}}\right]_{j}  \\
	& =\sqrt{w_2}\left[\frac{\partial{\int\mathrm I(E_S)\left[\int\mathrm h(r_{D'},r,E')S(r,\theta,E_S)f(r,r_S,E_S)\delta_{r_D,r_S}(r)\rho_n(r) \mathrm{d}r\right]\,\mathrm{d}E_S}}{\partial{p}}\right]_{j} \\
	 & =\sqrt{w_2}\left[\int\mathrm I(E_S)\left[\int \frac{\partial \left(h(r_{D'},r,E')f(r,r_S,E_S)\right)}{\partial p}S(r,\theta,E_S)\delta_{r_D,r_S}(r)\rho(r) \mathrm{d}r\right]\,\mathrm{d}E_S\right]_{j} 
\label{eq:49}
\end{split}
\end{equation}
where  $j \in {\{1,\dots,N_{CT}\}}$  indexes the number of rows in the forward model and Jacobian matrices.  The total number of scattered raypaths $N_{CT}= N_S\times N_D\times (N_D-1)\times N_E$, is defined by the number of sources and detectors, $N_S$ and $N_D$ and energy resolution of detectors, over which absorption data will be collected. 
Based on (\ref{eq:49}) the Jacobian matrix requires the computation of the first derivative of the attenuation coefficients for each broken raypath as 

\begin{equation}
\begin{split}
	 \frac{\partial \{h(r_{D'},r,E')f(r,r_S,E_S)\}}{\partial p} & =  \frac{\partial \{\Omega_{D'} \exp(-\int \mu(r'',E')\delta_{r_{D'},r}(r'') \mathrm{d}r'' -\int \mu(r'',E_S)\delta_{r,r_S}(r'') \mathrm{d}r'')\}}{\partial p} \\
	 &=(-E'^{-3}\int \delta_{r_{D'},r}(r'') \mathrm{d}r'' - E_S^{-3}\int \delta_{r,r_S}(r'') \mathrm{d}r'')) \times \\
&\Omega_{D'} \exp(-\int \mu(r'',E')\delta_{r_{D'},r}(r'') \mathrm{d}r'' -\int \mu(r'',E_S)\delta_{r,r_S}(r'') \mathrm{d}r'').
\end{split}
\label{eq:50}
\end{equation}
\section*{Acknowledgment}
This material is based upon work supported by the U.S. Department of Homeland Security, Science and Technology Directorate, Office of University Programs, under Grant Award 2013-ST-061-ED0001. The views and conclusions contained in this document are those of the authors and should not be interpreted as necessarily representing the official policies, either expressed or implied, of the U.S. Department of Homeland Security.

\bibliographystyle{IEEEtran}
\bibliography{ref}

\begin{thebibliography}{10}
\providecommand{\url}[1]{#1}
\csname url@samestyle\endcsname
\providecommand{\newblock}{\relax}
\providecommand{\bibinfo}[2]{#2}
\providecommand{\BIBentrySTDinterwordspacing}{\spaceskip=0pt\relax}
\providecommand{\BIBentryALTinterwordstretchfactor}{4}
\providecommand{\BIBentryALTinterwordspacing}{\spaceskip=\fontdimen2\font plus
\BIBentryALTinterwordstretchfactor\fontdimen3\font minus
  \fontdimen4\font\relax}
\providecommand{\BIBforeignlanguage}[2]{{%
\expandafter\ifx\csname l@#1\endcsname\relax
\typeout{** WARNING: IEEEtran.bst: No hyphenation pattern has been}%
\typeout{** loaded for the language `#1'. Using the pattern for}%
\typeout{** the default language instead.}%
\else
\language=\csname l@#1\endcsname
\fi
#2}}
\providecommand{\BIBdecl}{\relax}
\BIBdecl

\bibitem{r15}
B.~H. Tracey and E.~L. Miller, ``Stabilizing dual-energy x-ray computed
  tomography reconstructions using patch-based regularization,'' \emph{Inverse
  Problems}, vol.~31, no.~10, 2015.

\bibitem{r1}
H.~P. Hiriyannaiah, ``X-ray computed tomography for medical imaging,''
  \emph{Signal Processing Magazine, IEEE}, vol. 14.2, pp. 42--59, 1997.

\bibitem{r2}
M.~P. Hentschel, K.~w.~Harbich, and A.~Lange, ``Nondestructive evaluation of
  single fibre debonding in composites by x-ray refraction,'' \emph{NDT and E
  International}, vol. 27.5, pp. 275--280, 1994.

\bibitem{r3}
F.~e.~a. Mees, ``"applications of x-ray computed tomography in the
  geosciences." geological society,'' \emph{London, Special Publications}, vol.
  215.1, pp. 1--6, 2003.

\bibitem{r4}
P.~e.~a. Jin, ``A model-based 3d multi-slice helical ct reconstruction
  algorithm for transportation security application,'' \emph{Second
  International Conference on Image Formation in X-Ray Computed Tomography Salt
  Lake City, Utah, US}, 2012.

\bibitem{r5}
P.~M. Shikhaliev, ``Energy-resolved computed tomography: first experimental
  results,'' \emph{Physics in Medicine and Biology}, vol.~53, no.~20, pp.
  595--613, 2008.

\bibitem{r6}
T.~R. C. e.~a. Johnson, ``Material differentiation by dual energy ct: initial
  experience,'' \emph{European Radiology}, vol.~17, no.~6, pp. 1510--1517,
  2007.

\bibitem{r7}
W.~D. Engler, P.;~Friedman, ``Review of dual-energy computed tomography
  techniques,'' \emph{Materials Evaluation}, vol.~48, pp. 623--629, 1990.

\bibitem{r8}
\BIBentryALTinterwordspacing
A.~Gorecki, A.~Brambilla, V.~Moulin, E.~Gaborieau, P.~Radisson, and L.~Verger,
  ``Comparing performances of a cdte x-ray spectroscopic detector and an x-ray
  dual-energy sandwich detector,'' \emph{J. Instrumentation}, vol.~8, no.~11,
  p. P11011, 2013. [Online]. Available:
  \url{http://stacks.iop.org/1748-0221/8/i=11/a=P11011}
\BIBentrySTDinterwordspacing

\bibitem{r9}
P.~M. Shikhaliev and S.~G. Fritz, ``Photon counting spectral ct versus
  conventional ct: comparative evaluation for breast imaging application,''
  \emph{Physics in medicine and biology}, vol.~56, no.~7, p. 1905, 2011.

\bibitem{bushberg2011essential}
J.~T. Bushberg and J.~M. Boone, \emph{The essential physics of medical
  imaging}.\hskip 1em plus 0.5em minus 0.4em\relax Lippincott Williams \&
  Wilkins, 2011.

\bibitem{torikoshi2003electron}
M.~Torikoshi, T.~Tsunoo, M.~Sasaki, M.~Endo, Y.~Noda, Y.~Ohno, T.~Kohno,
  K.~Hyodo, K.~Uesugi, and N.~Yagi, ``Electron density measurement with
  dual-energy x-ray ct using synchrotron radiation,'' \emph{Physics in medicine
  and biology}, vol.~48, no.~5, p. 673, 2003.

\bibitem{r10}
Y.~Zhang, X.~Mou, G.~Wang, and H.~Yu, ``Tensor-based dictionary learning for
  spectral ct reconstruction,'' \emph{IEEE Transactions on Medical Imaging},
  2016.

\bibitem{r11}
O.~Semerci, N.~Hao, M.~E. Kilmer, and E.~L. Miller, ``Tensor-based formulation
  and nuclear norm regularization for multienergy computed tomography,''
  \emph{IEEE Transactions on Image Processing}, vol.~23, no.~4, pp. 1678--1693,
  2014.

\bibitem{r12}
M.~Wang, Y.~Zhang, R.~Liu, S.~Guo, and H.~Yu, ``An adaptive reconstruction
  algorithm for spectral ct regularized by a reference image,'' \emph{Physics
  in Medicine and Biology}, vol.~61, no.~24, p. 8699, 2016.

\bibitem{r13}
A.~H. Andersen and A.~C. Kak, ``Simultaneous algebraic reconstruction technique
  (sart): a superior implementation of the art algorithm,'' \emph{Ultrasonic
  imaging}, vol.~6, no.~1, pp. 81--94, 1984.

\bibitem{r14}
O.~Semerci and E.~L. Miller, ``A parametric level-set approach to simultaneous
  object identification and background reconstruction for dual-energy computed
  tomography,'' \emph{Image Processing, IEEE Transactions on}, vol. 21.5, pp.
  2719--2734, 2012.

\bibitem{r16}
Y.~Zhang, Y.~Xi, Q.~Yang, W.~Cong, J.~Zhou, and G.~Wang, ``Spectral ct
  reconstruction with image sparsity and spectral mean,'' \emph{IEEE
  Transactions on Computational Imaging}, vol.~2, no.~4, 2016.

\bibitem{r17}
Y.~Wang, G.~Wang, S.~Mao, W.~Cong, Z.~Ji, J.-F. Cai, and Y.~Ye, ``A
  framelet-based iterative maximum-likelihood reconstruction algorithm for
  spectral ct,'' \emph{Inverse Problems}, vol.~32, no.~11, p. 115021, 2016.

\bibitem{r18}
K.~Kim, J.~C. Ye, W.~Worstell, J.~Ouyang, Y.~Rakvongthai, G.~El~Fakhri, and
  Q.~Li, ``Sparse-view spectral ct reconstruction using spectral patch-based
  low-rank penalty,'' \emph{IEEE transactions on medical imaging}, vol.~34,
  no.~3, pp. 748--760, 2015.

\bibitem{r19}
Y.~S. Han, K.~H. Jin, K.~Kim, and J.~C. Ye, ``Sparse-view x-ray spectral ct
  reconstruction using annihilating filter-based low rank hankel matrix
  approach,'' in \emph{Biomedical Imaging (ISBI), 2016 IEEE 13th International
  Symposium on}.\hskip 1em plus 0.5em minus 0.4em\relax IEEE, 2016, pp.
  573--576.

\bibitem{r20}
A.~P. Yazdanpanah, E.~E. Regentova, and G.~Bebis, ``Algebraic iterative
  reconstruction-reprojection (airr) method for high performance sparse-view ct
  reconstruction,'' \emph{Appl. Math}, vol.~10, no.~6, pp. 1--8, 2016.

\bibitem{r21}
S.~J. Norton, ``Compton scattering tomography,'' \emph{Journal of applied
  physics}, vol. 76.4, pp. 2007--2015, 1994.

\bibitem{r22}
e.~a. Lange, Axel, ``X-ray compton tomography,'' \emph{11th European Conference
  on Non-Destructive Testing (ECNDT 2014)October 6-10, Prague, Czech Republic},
  vol. 21.5, 2014.

\bibitem{pfeiffer2008hard}
F.~Pfeiffer, M.~Bech, O.~Bunk, P.~Kraft, E.~F. Eikenberry, C.~Br{\"o}nnimann,
  C.~Gr{\"u}nzweig, and C.~David, ``Hard-x-ray dark-field imaging using a
  grating interferometer,'' \emph{Nature materials}, vol.~7, no.~2, pp.
  134--137, 2008.

\bibitem{r32}
W.cong and G.Wang, ``X-ray scattering tomography for biological applications,''
  \emph{Journal of X-Ray Science and Technology}, vol.~19, no.~2, pp. 219--227,
  2011.

\bibitem{r23}
T.~T. Truong and M.~K. Nguyen, \emph{Recent Developments on Compton Scatter
  Tomography: Theory and Numerical Simulations}.\hskip 1em plus 0.5em minus
  0.4em\relax INTECH Open Access, 2012.

\bibitem{r24}
N.~Kondic, A.~Jacobs, and D.~Ebert, \emph{Three-dimensional density field
  determination by external stationary detectors and gamma sources using
  selective scattering}.\hskip 1em plus 0.5em minus 0.4em\relax Thermal
  hydraulics of nuclear reactors, 1983.

\bibitem{r25}
M.~K. Nguyen and T.~T. Truong, ``Inversion of a new circular-arc radon
  transform for compton scattering tomography,'' \emph{Inverse Problems},
  vol.~26, no. 065005, p.~6, 2010.

\bibitem{r26}
M.~K. e.~a. Nguyen, \emph{"A novel technological imaging process using ionizing
  radiation properties." Computing and Communication Technologies, Research,
  Innovation, and Vision for the Future (RIVF), 2012 IEEE RIVF International
  Conference on}.\hskip 1em plus 0.5em minus 0.4em\relax IEEE, 2012.

\bibitem{r27}
J.~Webber, ``X-ray compton scattering tomography,'' vol.~6, May 2015.

\bibitem{r28}
Z.~C. Jiajun~Wang and Y.~Wang, ``Analytic reconstruction of compton scattering
  tomography,'' \emph{Journal of Applied Physics}, vol.~86, no.~3, pp.
  1693--1698, 1999.

\bibitem{r29}
F.~Zhao, J.~C. Schotland, and V.~A. Markel, ``Inversion of the star
  transform,'' \emph{Inverse Problems}, vol.~30, no. 105001, p.~10, 2014.

\bibitem{r30}
R.~Krylov and A.~Katsevich, ``Inversion of the broken ray transform in the case
  of energy-dependent attenuation,'' \emph{Physics in Medicine and Biology},
  vol.~60, no. 4313, p.~11, 2015.

\bibitem{r31}
B.~e.~a. Golosio, ``Internal elemental microanalysis combining x-ray
  fluorescence, compton and transmission tomography,'' \emph{Journal of applied
  Physics}, vol.~94, no.~1, pp. 145--156, 2003.

\bibitem{r36}
O.~Semerci, ``Image formation methods for dual energy and multi-energy computed
  tomography,'' Ph.D. dissertation, October 2012.

\bibitem{alvarez1976energy}
R.~E. Alvarez and A.~Macovski, ``Energy-selective reconstructions in x-ray
  computerised tomography,'' \emph{Physics in medicine and biology}, vol.~21,
  no.~5, p. 733, 1976.

\bibitem{evans1955atomic}
R.~D. Evans and A.~Noyau, \emph{The atomic nucleus}.\hskip 1em plus 0.5em minus
  0.4em\relax McGraw-Hill New York, 1955, vol. 582.

\bibitem{hartemann2001three}
F.~V. Hartemann, B.~Rupp, H.~Baldis, D.~Gibson, A.~Kerman, and A.~Le~Foll,
  ``Three-dimensional theory of compton scattering and advanced biomedical
  applications,'' in \emph{Particle Accelerator Conference, 2001. PAC 2001.
  Proceedings of the 2001}, vol.~4.\hskip 1em plus 0.5em minus 0.4em\relax
  IEEE, 2001, pp. 2641--2643.

\bibitem{r35}
N.~Zaluzec, ``Analytical formulae for calculation of x-ray detector solid
  angles in the scanning and scanning/transmission analytical electron
  microscope,'' \emph{Microscopy and Microanalysis}, vol.~20, no.~4, p.
  1318–1326, 2014.

\bibitem{r39}
S.~J. Norton, ``Compton scattering tomography,'' \emph{Journal of applied
  physics}, vol.~76, no.~4, pp. 2007--2015, 1994.

\bibitem{sonka2000handbook}
M.~Sonka and J.~M. Fitzpatrick, ``Handbook of medical imaging(volume 2, medical
  image processing and analysis).''\hskip 1em plus 0.5em minus 0.4em\relax
  SPIE- The international society for optical engineering, 2000.

\bibitem{tang2009performance}
J.~Tang, B.~E. Nett, and G.-H. Chen, ``Performance comparison between total
  variation (tv)-based compressed sensing and statistical iterative
  reconstruction algorithms,'' \emph{Physics in medicine and biology}, vol.~54,
  no.~19, p. 5781, 2009.

\bibitem{siltanen2003statistical}
S.~Siltanen, V.~Kolehmainen, S.~J{\"a}rvenp{\"a}{\"a}, J.~Kaipio, P.~Koistinen,
  M.~Lassas, J.~Pirttil{\"a}, and E.~Somersalo, ``Statistical inversion for
  medical x-ray tomography with few radiographs: I. general theory,''
  \emph{Physics in medicine and biology}, vol.~48, no.~10, p. 1437, 2003.

\bibitem{jain2005score}
A.~Jain, K.~Nandakumar, and A.~Ross, ``Score normalization in multimodal
  biometric systems,'' \emph{Pattern recognition}, vol.~38, no.~12, pp.
  2270--2285, 2005.

\bibitem{bouman1996unified}
C.~A. Bouman and K.~Sauer, ``A unified approach to statistical tomography using
  coordinate descent optimization,'' \emph{IEEE Transactions on image
  processing}, vol.~5, no.~3, pp. 480--492, 1996.

\bibitem{fan2010maximum}
W.~Fan and H.~Wang, ``Maximum entropy regularization method for electrical
  impedance tomography combined with a normalized sensitivity map,'' \emph{Flow
  Measurement and Instrumentation}, vol.~21, no.~3, pp. 277--283, 2010.

\bibitem{muniz2000entropy}
W.~Muniz, F.~Ramos, and H.~de~Campos~Velho, ``Entropy-and tikhonov-based
  regularization techniques applied to the backwards heat equation,''
  \emph{Computers \& mathematics with Applications}, vol.~40, no.~8, pp.
  1071--1084, 2000.

\bibitem{xu2003minimum}
X.~Xu, E.~L. Miller, and C.~M. Rappaport, ``Minimum entropy regularization in
  frequency-wavenumber migration to localize subsurface objects,'' \emph{IEEE
  Transactions on Geoscience and Remote Sensing}, vol.~41, no.~8, pp.
  1804--1812, 2003.

\bibitem{kivinen1997exponentiated}
J.~Kivinen and M.~K. Warmuth, ``Exponentiated gradient versus gradient descent
  for linear predictors,'' \emph{Information and Computation}, vol. 132, no.~1,
  pp. 1--63, 1997.

\bibitem{burger2016bregman}
M.~Burger, ``Bregman distances in inverse problems and partial differential
  equations,'' in \emph{Advances in Mathematical Modeling, Optimization and
  Optimal Control}.\hskip 1em plus 0.5em minus 0.4em\relax Springer, 2016, pp.
  3--33.

\bibitem{r33}
C.~C. Paige and M.~A. Saunders, ``Lsqr: An algorithm for sparse linear
  equations and sparse least squares,'' \emph{ACM Transactions on Mathematical
  Software (TOMS)}, vol.~8, no.~1, pp. 43--71, 2003.

\bibitem{r34}
D.~Marquardt, ``An algorithm for least-squares estimation of nonlinear
  parameters,'' \emph{Journal of the Society for Industrial And Applied
  Mathematics}, vol.~11, no.~2, pp. 145--156, 1963.

\bibitem{buades2005non}
A.~Buades, B.~Coll, and J.-M. Morel, ``A non-local algorithm for image
  denoising,'' in \emph{Computer Vision and Pattern Recognition, 2005. CVPR
  2005. IEEE Computer Society Conference on}, vol.~2.\hskip 1em plus 0.5em
  minus 0.4em\relax IEEE, 2005, pp. 60--65.

\bibitem{darbon2008fast}
J.~Darbon, A.~Cunha, T.~F. Chan, S.~Osher, and G.~J. Jensen, ``Fast nonlocal
  filtering applied to electron cryomicroscopy,'' in \emph{Biomedical Imaging:
  From Nano to Macro, 2008. ISBI 2008. 5th IEEE International Symposium
  on}.\hskip 1em plus 0.5em minus 0.4em\relax IEEE, 2008, pp. 1331--1334.

\bibitem{xcom}
M.~Berger, J.~Hubbell, S.~Seltzer, J.~Chang, J.~Coursey, R.~Sukumar, and
  D.~Zucker, ``Xcom: Photon cross sections database,'' \emph{NIST Standard
  Reference Database}, vol.~8, no.~6, p. 87–3597, 1998.

\bibitem{r37}
C.~Vogel, \emph{Computational Methods for Inverse Problems}.\hskip 1em plus
  0.5em minus 0.4em\relax Society for Industrial and Applied Mathematics, 2002.

\bibitem{K.Madsen2004}
K.~Madsen, H.~Nielsen, and O.~Tingleff, ``Methods for non-linear least squares
  problems,'' \emph{Technical University of Denmark}, 2004.

\end{thebibliography}


\end{document}